\documentclass[10pt,journal,compsoc]{IEEEtran}
\usepackage[nocompress]{cite}
\usepackage{times}
\usepackage{soul}
\usepackage{url}
\usepackage{makecell}
\usepackage[hidelinks]{hyperref}
\usepackage[utf8]{inputenc}
\usepackage[small]{caption}   
\usepackage{graphicx}
\usepackage{amsthm}
\usepackage{booktabs}
\usepackage{algorithm}
\usepackage{algorithmic}
\usepackage{array}
\usepackage{threeparttable}
\usepackage{tabularx}
\usepackage{enumerate}
\usepackage{bbm}
\usepackage{epsfig, epstopdf}
\usepackage{multirow}
\usepackage{xcolor}

\usepackage{flushend}
\usepackage{subfigure}
\usepackage{amsmath,amsfonts,amssymb}
\usepackage{lineno}
\usepackage{siunitx}
\usepackage{CJKutf8}

\usepackage{indentfirst}
\usepackage[T1]{fontenc}
\usepackage{ragged2e}
\renewcommand{\raggedright}{\leftskip=0pt \rightskip=0pt plus 0cm}

\ifCLASSOPTIONcompsoc
  \usepackage[nocompress]{cite}
\else
  \usepackage{cite}
\fi
\ifCLASSINFOpdf
\else
\fi
\begin{document}\sloppy
\title{Deep Clustering: A Comprehensive Survey}

\author{Yazhou Ren,~\IEEEmembership{Member,~IEEE}, Jingyu Pu, Zhimeng Yang, Jie Xu, Guofeng Li, Xiaorong Pu, \\Philip S. Yu,~\IEEEmembership{Fellow,~IEEE}, Lifang He,~\IEEEmembership{Member,~IEEE}
\IEEEcompsocitemizethanks{\IEEEcompsocthanksitem 
Yazhou Ren, Jingyu Pu, Zhimeng Yang, Jie Xu, Guofeng Li and Xiaorong Pu are with University of Electronic Science and Technology of China, Chengdu 611731, China. Yazhou Ren is the corresponding author. E-mail: yazhou.ren@uestc.edu.cn.
\IEEEcompsocthanksitem Philip S. Yu is with University of Illinois at Chicago, IL 60607, USA.
\IEEEcompsocthanksitem Lifang He is with Lehigh University, PA 18015, USA.}
\thanks{Manuscript received Oct. 2022.}
}

\IEEEtitleabstractindextext{%
\begin{abstract}
Cluster analysis plays an indispensable role in machine learning and data mining. Learning a good data representation is crucial for clustering algorithms. Recently, deep clustering, which can learn clustering-friendly representations using deep neural networks, has been broadly applied in a wide range of clustering tasks. Existing surveys for deep clustering mainly focus on the single-view fields and the network architectures, ignoring the complex application scenarios of clustering. To address this issue, in this paper we provide a comprehensive survey for deep clustering in views of data sources. With different data sources and initial conditions, we systematically distinguish the clustering methods in terms of methodology, prior knowledge, and architecture.
Concretely, deep clustering methods are introduced according to four categories, i.e., traditional single-view deep clustering, semi-supervised deep clustering, deep multi-view clustering, and deep transfer clustering. Finally, we discuss the open challenges and potential future opportunities in different fields of deep clustering.

\end{abstract}

}

\maketitle

\IEEEdisplaynontitleabstractindextext

%
\IEEEpeerreviewmaketitle

\IEEEraisesectionheading{\section{Introduction}\label{sec:introduction}}

%
%
%
%
\IEEEPARstart{W}{ith} the development of online media, abundant data with high complexity can be gathered easily. Through pinpoint analysis of these data, we can dig the value out and use these conclusions in many fields, such as face recognition \cite{wang2020masked,guo2020learning}, sentiment analysis \cite{yadav2020sentiment,xu2019sentiment}, intelligent manufacturing \cite{zhou2018toward,zhou2019human}, etc.


A model which can be used to classify the data with different labels is the base of many applications. For labeled data, it is taken granted to use the labels as the most important information as a guide. For unlabeled data, finding a quantifiable objective as the guide of the model-building process is the key question of clustering.
Over the past decades, a large number of clustering methods with shallow models have been proposed, including centroid-based clustering \cite{MACQUEEN1967SomeMF,Ren2019PBC}, density-based clustering \cite{ester1996density,comaniciu2002mean,Ren2014Boosted,Ren2014weighted,ren2018semi}, distribution-based clustering \cite{Bishop2006Pattern_SR}, hierarchical clustering \cite{Jain:clustering}, ensemble clustering \cite{Strehl2002cluster,Ren2017Weighted}, multi-view clustering \cite{kumar2011co,kumar2011coreg,cai2013multi,huang2021non,huang2021dual,huang2018robust}, etc. These shallow models are effective only when the features are representative, while their performance on the complex data is usually limited due to the poor power of feature learning.

\par In order to map the original complex data to a feature space that is easy to cluster, many clustering methods focus on feature extraction or feature transformation, such as PCA \cite{wold1987principal}, kernel method \cite{hearst1998support}, spectral method \cite{feit1982solution}, deep neural network \cite{liu2017survey}, etc. Among these methods, the deep neural network is a promising approach because of its excellent nonlinear mapping capability and its flexibility in different scenarios. A well-designed deep learning based clustering approach (referred to deep clustering) aims at effectively extracting more clustering-friendly features from data and performing clustering with learned features simultaneously.



\par Much research has been done in the field of deep clustering and there are also some surveys about deep clustering methods  \cite{aljalbout2018clustering, min2018survey,nutakki2019introduction,zhou2022comprehensive}. 
Specifically, existing systematic reviews for deep clustering mainly focus on the single-view clustering tasks and the architectures of neural networks.
For example, Aljalbout \emph{et al}. \cite{aljalbout2018clustering} focus only on deep single-view  clustering methods which are based on deep autoencoder (AE or DAE). Min \emph{et al}. \cite{min2018survey} classify deep clustering methods from the perspective of different deep networks. Nutakki \emph{et al}. \cite{nutakki2019introduction} divide deep single-view clustering methods into three categories according to their training strategies: multi-step sequential deep clustering, joint deep clustering, and closed-loop multi-step deep clustering.
Zhou \emph{et al.} \cite{zhou2022comprehensive} categorize deep single-view clustering methods by the interaction way between feature learning and clustering modules.
But in the real world, the datasets for clustering are always associated, e.g., the taste for reading is correlated with the taste for a movie, and the side face and full-face from the same person should be labeled the same. For these data, deep clustering methods based on semi-supervised learning, multi-view learning, and transfer learning have also made significant progress. Unfortunately, existing reviews do not discuss them too much. 

\begin{figure}[!t]
\centering
\includegraphics[scale=0.3]{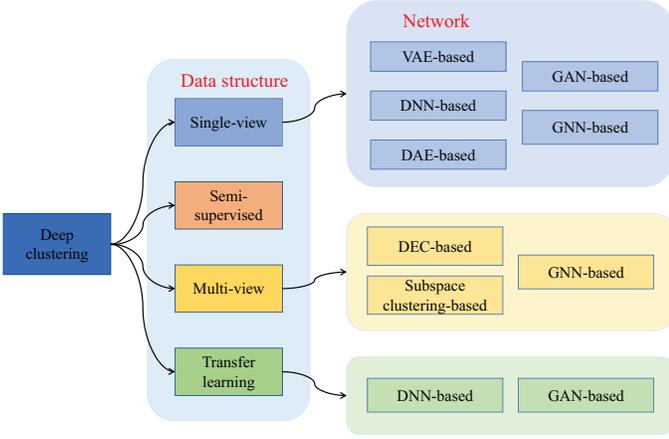}
\caption{The directory tree of this survey.}
\label{fig:survey-tree}
\end{figure}

Therefore, it is important to classify deep clustering from the perspective of data sources and initial conditions.
In this survey, we summarize the deep clustering from the perspective of initial settings of data combined with deep learning methodology. We introduce the newest progress of deep clustering from the perspective of network and data structure as shown in Fig.~\ref{fig:survey-tree}. Specifically, we organize the deep clustering methods into the following four categories:

\begin{itemize}
    \item \textbf{Deep single-view clustering}
\end{itemize}

For conventional clustering tasks, it is often assumed that the data are of the same form and structure, as known as single-view or single-modal data.
The extraction of representations for these data by deep neural networks (DNNs) is a significant characteristic of deep clustering. However, what is more noteworthy is the different applied deep learning techniques, which are highly correlated with the structure of DNNs. To compare the technical route of specific DNNs, we divide those algorithms into five categories: deep autoencoder (DAE) based deep clustering, 
deep neural network (DNN) based deep clustering, variational autoencoder (VAE) based deep clustering, generative adversarial network (GAN) based deep clustering and graph nerual network (GNN) based deep clustering. 

\begin{itemize}
    \item \textbf{Deep clustering based on semi-supervised learning}
\end{itemize}

When the data to be processed contain a small part of prior constraints, 
traditional clustering methods cannot effectively utilize this prior information and semi-supervised clustering is an effective way to solve this question. In presence, the research of deep semi-supervised clustering has not been well explored. 
However, semi-supervised clustering is inevitable because it is feasible to let a clustering method become a semi-supervised one by adding the additional information as a constraint loss to the model.

\begin{itemize}
    \item \textbf{Deep clustering based on multi-view learning}
\end{itemize}

In the real world, data are often obtained from different feature collectors or have different structures. We call those data ''multi-view data'' or ''multi-modal data'', where each sample has multiple representations. The purpose of deep clustering based on multi-view learning is to utilize the consistent and complementary information contained in multi-view data to improve clustering performance. In addition, the idea of multi-view learning may have guiding significance for deep single-view clustering. In this survey, we summarize deep multi-view clustering into three categories: deep embedded clustering based, subspace clustering based, and graph neural network based.

\begin{itemize}
    \item \textbf{Deep clustering based on transfer learning}
\end{itemize}

For a task that has a limited amount of instances and high dimensions, sometimes we can find an assistant to offer additional information. For example, if task A is similar to another task B and B has more information for clustering than A (B is labeled or B is easier to  clustering than A), it is useful to transfer the information from B to A. Transfer learning for unsupervised domain adaption (UDA) is boosted in recent years, which contains two domains: Source domain with labels and target domain which is unlabeled. The goal of transfer learning is to apply the knowledge or patterns learned from the source task to a different but related target task. Deep clustering methods based on transfer learning aim to improve the performance of current clustering tasks by utilizing information from relevant tasks.

It is necessary to pay attention to the different characteristics and conditions of the clustering data before studying the corresponding clustering methods. In this survey,  existing deep clustering methods are systematically classified from data sources and initial conditions. 
The advantages, disadvantages, and applicable conditions of different clustering methods are analyzed. Finally, we present some interesting research directions in the field of deep clustering.

\section{Definitions and Preliminaries}
We introduce the notations in this section. Throughout this paper, we use uppercase letters to denote matrices and lowercase letters to denote vectors. Unless otherwise stated, the notations used in this paper are summarized in Table \ref{tab: Notations}.
\begin{table}[!t]
  \centering
  \caption{ Notations and their descriptions in this paper.
}
    \begin{tabular}{|c|l|}
    \hline
    \multicolumn{1}{|l|}{Notations} & \multicolumn{1}{c|}{Descriptions} \\ \hline
    $i$  & a counter variable \\ \hline
    $j$  & a counter variable \\ \hline
    $|.|$ & the length of a set \\ \hline
    $\lVert . \rVert $ & the 2-norm of a vector \\ \hline
    $X$ & the data for clustering \\ \hline
    $X^s$ & the data in source domain (UDA methods)\\ \hline
    $Y^s$ & the labels of source domain instances (UDA methods)\\ \hline
    $X^t$ & the data in target domain (UDA methods)\\ \hline
    $\mathcal{D}_s$ & the source domain of UDA methods \\ \hline
    $\mathcal{D}_t$ & the target domain of UDA methods \\ \hline
    $x_i$ & the vector of an oringinal data sample \\ \hline
    $X^i$ & the $i$-th view of $X$ in multi-view learning \\ \hline
    $\hat{Y} $& the predicted labels of $X$ \\ \hline
    $S$ & the soft data assignments of $X$ \\ \hline
    $R$ & the adjusted assignments of $S$ \\ \hline
    $A$ & the pairwise constraint matrix\\ \hline
    $a_{ij}$ & the constraint of sample $i$ and sample $j$\\ \hline
    $z_i$ & the vector of the embedded representation of $x_i$\\ \hline
    $\varepsilon$ & the noise used in generative model \\ \hline
    $\mathbb{E}$ & the expectation \\ \hline
    $L_n$ & the network loss \\ \hline
    $L_c$ & the clustering loss \\ \hline
    $L_{ext}$ & the extra task loss \\ \hline
    $L_{rec}$ & the reconstruction loss of autoencoder network \\ \hline
    $L_{gan}$ & the loss of GAN \\ \hline
    $L_{\scriptscriptstyle ELBO}$ & the loss of evidence lower bound \\ \hline
    $k$ & the number of clusters  \\ \hline
    $n$ & the number of data samples \\ \hline
    $\mu$ & the mean of the Gaussian distribution \\ \hline
    $\theta$ & the variance of the Gaussian distribution \\ \hline
    $KL(. \Vert .)$ & the Kullback-Leibler divergence \\ \hline 
    $p(.)$ & the probability distribution \\ \hline 
    $p(.|.)$ & the conditional probability distribution\\ \hline 
    $p(.,.)$ & the joint probability distribution\\ \hline 
    $q(.)$ & the approximate probability distribution of $p(.)$ \\ \hline 
    $q(.|.)$ & the approximate probability distribution of $p(.|.)$\\ \hline 
    $q(.,.)$ & the approximate probability distribution of $p(.,.)$\\ \hline 
    $f(.)$ & the feature extractor\\ \hline 
    ${\phi}_e(.)$ & the encoder network of AE or VAE \\ \hline 
    ${\phi}_r(.)$ & the decoder network of AE or VAE \\ \hline 
    ${\phi}_g(.)$ & the generative network of GAN \\ \hline 
    ${\phi}_d(.)$ & the discriminative network of GAN \\ \hline 
    $Q$ & the graph adjacency matrix  \\ \hline
    $D$ & the degree matrix of  $Q$ \\ \hline
    $C$ & the feature matrix of a graph \\ \hline
    $H$ & the node hidden feature matrix  \\ \hline
    $W$ & the learnable model parameters  \\ \hline
    \end{tabular}%
  \label{tab: Notations}%
\end{table}%

\par This survey will introduce four kinds of deep clustering problems based on different background conditions. Here, we define these problems formally. Given a set of data samples $X$, we aim at finding a map function $F$ which can map $X$ into $k$ clusters. The map result is represented with $\hat{Y}$. So the tasks we cope with are:

(1) Deep single-view clustering: 
\begin{equation}
\label{Single-view deep clustering task}
F\left( X \right) \rightarrow \hat{Y}.
\end{equation}

(2) Semi-supervised deep clustering: 
\begin{equation}
\label{Semi-supervised deep clustering task}
F\left( X,A \right) \rightarrow \hat{Y},
\end{equation}
 where $A$ is  a constrained matrix.

(3) Deep multi-view clustering: 
\begin{equation}
\label{Multi-view deep clustering task}
F\left( X^1,...,X^n \right) \rightarrow \hat{Y},
\end{equation}
where $X^i$ is the $i$-th view of $X$.

(4) Deep clustering with domain adaptation: 
\begin{equation}
\label{Unsupervised domain adaptation task}
F\left( X^s,Y^s,X^t \right) \rightarrow \hat{Y},
\end{equation}
where $(X^s,Y^s)$ is the labeled source domain and $X^t$ is the unlabeled target domain.

\section{Deep Single-view Clustering}
The theory of representation learning \cite{Yoshua2013Representation} shows the importance of feature learning (or representation learning) in machine learning tasks.
However, deep representation learning is mostly supervised learning that requires many labeled data.
As we mentioned before, the obstacle of the deep clustering problem is what can be used to guide the training process like labels in supervised problem.
The most ``supervised'' information in deep clustering is the data itself. So how can we train an effective feature extractor to get good representation? According to the way the feature extractor is trained, we divide deep single-view clustering algorithms into five categories: \emph{DAE-based}, \emph{DNN-based},  \emph{VAE-based}, \emph{GAN-based},  and \emph{GNN-based}. The difference of these methods is mainly about the loss components, where the loss terms are defined in Table 1 and explained below:
\begin{itemize} 
\item
\emph{DAE-based}/\emph{GNN-based}: $L = L_{rec}+L_c$, 
\item
\emph{DNN-based}: $L = L_{ext}+L_c$, 
\item
\emph{VAE-based}: $L = L_{\scriptscriptstyle ELBO}+L_c$, 
\item
\emph{GAN-based}: $L = L_{gan}+L_c$. 
\end{itemize}

In unsupervised learning, the issue we cope with is to train a reliable feature extractor without labels. There are mainly two ways in existing works:
1) A loss function that optimizes the pseudo labels according to the principle: narrowing the inner-cluster distance and widening the inter-cluster distance. 
2) An extra task that can help train the feature extractor.
For the clustering methods with specialized feature extractors, such as autoencoder, the reconstruction loss $L_{rec}$ can be interpreted as the extra task.
In this paper, the clustering-oriented loss $L_c$ indicates the loss of the clustering objective. \emph{DAE-based}/\emph{GNN-based} methods use an autoencoder/graph autoencoder as the feature extractor, so the loss functions are always composed of a reconstruction loss $L_{rec}$ and another clustering-oriented loss $L_c$. By contrast, \emph{DNN-based} methods optimize the feature extractor with extra tasks or other strategies $L_{ext}$. \emph{VAE-based} methods optimize the loss of evidence lower bound $L_{\scriptscriptstyle ELBO}$. \emph{GAN-based} methods are based on the generative adversarial loss $L_{gan}$.
Based on these five dimensions, existing deep single-view clustering methods are summarized in Table \ref{The summaries of DAE-based and DNN-based methods in single-view clustering.} and Table \ref{The summaries of VAE-based and GAN-based methods in single-view clustering.}. 

\subsection{DAE-based}\label{sec:DAE-based}
\begin{table*}[!t]
  \centering
  \caption{The summaries of \emph{DAE-based} and \emph{DNN-based} methods in deep single-view clustering. We summarize the \emph{DAE-based} methods based on ``Jointly or Separately'' and  ``Characteristics''.}
    \begin{tabular}{|c|r|c|l|}
 
    \hline
    Net   & \multicolumn{1}{c|}{Methods} &\multicolumn{1}{m{2cm}|}{Jointly or Separately} & \multicolumn{1}{c|}{Characteristics} \\
    \hline
    \multirow{19}[20]*{DAE} & AEC (2013) \cite{song2013auto} & Separately & Optimize the distance between $z_i$ and its closest cluster centroid. \\
\cline{2-4}          & DEN (2014) \cite{huang2014deep} & Separately & Locality-preserving constraint, group sparsity constraint. \\
\cline{2-4}          & PARTY (2016) \cite{peng2016deep} & Separately & Subspace clustering. \\
\cline{2-4}          & DEC (2016) \cite{xie2016unsupervised} & Jointly  & Optimize the distribution of assignments. \\
\cline{2-4}          & IDEC  (2017) \cite{guo2017improved} & Jointly  & Improve DEC \cite{xie2016unsupervised} with local structure preservation.  \\
\cline{2-4}          & DSC-Nets (2017) \cite{ji2017deep} & Separately & Subspace clustering. \\
\cline{2-4}          & DEPICT (2017) \cite{ghasedi2017deep} & Jointly  &  Convolutional autoencoder and relative entropy minimization.\\ 
\cline{2-4}          & DCN (2017) \cite{yang2017towards} & Jointly  & Take the objective of $k$-means as the clustering loss. \\
\cline{2-4}          & DMC (2017) \cite{chen2017unsupervised} & Jointly  & Multi-manifold clustering. \\
\cline{2-4}          & DEC-DA  (2018) \cite{guo2018deep} & Jointly  & Improve DEC \cite{xie2016unsupervised} with data augmentation. \\
\cline{2-4}          & DBC (2018) \cite{li2018discriminatively} & Jointly  & Self-paced learning. \\
\cline{2-4}          & DCC (2018) \cite{shah2018deep} & Separately & Extend robust continuous clustering \cite{shah2017robust} with autoencoder. Not given $k$. \\
\cline{2-4}          & DDLSC (2018) \cite{tzoreff2018deep} & Jointly  & Pairwise loss function. \\
\cline{2-4}          &  DDC (2019) \cite{chang2019deep} & Separately & Global and local constraints of relationships. \\
\cline{2-4}          & DSCDAE (2019) \cite{yang2019deep} & Jointly  & Subspace Clustering. \\
\cline{2-4}          & NCSC (2019) \cite{zhang2019neural} & Jointly  & Dual autoencoder network. \\
\cline{2-4}          & DDIC (2020) \cite{ren2020deep} & Separately & Density-based clustering. Not given $k$. \\
\cline{2-4}          & SC-EDAE (2020) \cite{affeldt2020spectral}& Jointly  & Spectral clustering. \\
\cline{2-4}          & ASPC-DA (2020) \cite{guo2020adaptive}& Jointly  & Self-paced learning and data augmentation. \\
\cline{2-4}          & ALRDC (2020) \cite{yang2020adversarial}& Jointly  & Adversarial learning. \\
\cline{2-4}          & N2D (2021) \cite{mcconville2021n2d}& Separately  & Manifold learning. \\ 
\cline{2-4}          & AGMDC (2021) \cite{wang2021unsupervised}& Jointly& Gaussian Mixture Model. Improve the inter-cluster distance. \\ 
   \hline
    Net   & \multicolumn{1}{c|}{Methods} & \multicolumn{1}{m{2cm}|}{Clustering-oriented loss} & \multicolumn{1}{c|}{Characteristics} \\
    \hline
    \multirow{7}[15]{*}{DNN} & JULE (2016) \cite{yang2016joint} & Yes   & Agglomerative clustering. \\
\cline{2-4}          & DDBC (2017) \cite{kampffmeyer2017deep} & Yes   & Information theoretic measures. \\
\cline{2-4}          & DAC (2017) \cite{chang2017deep} & No    & Self-adaptation learning. Binary pairwise-classification. \\
\cline{2-4}          & DeepCluster (2018) \cite{caron2018deep} & No    & Use traditional clustering methods to assign labels. \\
\cline{2-4}          & CCNN (2018) \cite{hsu2018cnn} & No    & Mini-batch $k$-means. Feature drift compensation for large-scale image data \\
\cline{2-4}          & ADC (2018) \cite{haeusser2018associative} & Yes   & Centroid embeddings. \\
\cline{2-4}          & ST-DAC (2019) \cite{souza2019improving} & No    & Spatial transformer layers. Binary pairwise-classification. \\
\cline{2-4}          & RTM (2019) \cite{nina2019decoder} & No    & Random triplet mining. \\
\cline{2-4}          & IIC (2019) \cite{ji2019invariant} & No    & Mutual information. Generated image pairs. \\
\cline{2-4}          & DCCM (2019) \cite{wu2019deep} & No    & Triplet mutual information. Generated image pairs. \\
\cline{2-4}          & MMDC (2019) \cite{shiran2019multi} & No    & Multi-modal. Generated image pairs. \\
\cline{2-4}          & SCAN (2020) \cite{wvangansbeke2020scan} & Yes   & Decouple feature learning and clustering. Nearest neighbors  mining. \\
\cline{2-4}          & DRC (2020) \cite{zhong2020deep} & Yes   & Contrastive learning. \\
\cline{2-4}          & PICA (2020) \cite{huang2020deep} & Yes   & Maximize the “global” partition confidence.  \\
     \hline

    \end{tabular}%
    
  \label{The summaries of DAE-based and DNN-based methods in single-view clustering.}%
\end{table*}%

\begin{table*}[!t]
  \centering
  \caption{The summaries of \emph{VAE-}, \emph{GAN-}, and \emph{GNN-based} methods in deep single-view clustering.}
    \begin{tabular}{|c|r|c|l|}
    \hline

    \multicolumn{1}{|c|}{Net} & \multicolumn{1}{c|}{Methods} & \multicolumn{2}{c|}{Characteristics} \\
    \hline
    \multirow{3}[14]{*}{VAE} & VaDE (2016) \cite{jiang2016variational} &        \multicolumn{2}{l|}{Gaussian mixture variational autoencoder. } \\
\cline{2-4}          & GMVAE (2016) \cite{dilokthanakul2016deep} & \multicolumn{2}{l|}{ Gaussian mixture variational autoencoder. Unbalanced clustering.} \\
\cline{2-4}          & MFVDC (2017) \cite{figueroa2017simple} &      \multicolumn{2}{l|}{ Continuous Gumbel-Softmax distribution.} \\
\cline{2-4}          & LTVAE (2018) \cite{li2018learning} &    \multicolumn{2}{l|}{Latent tree model.} \\
\cline{2-4}          & VLAC (2019) \cite{willetts2019disentangling} &    \multicolumn{2}{l|}{ Variational ladder autoencoders.} \\
\cline{2-4}          & VAEIC (2020) \cite{prasad2020variational} &      \multicolumn{2}{l|}{ No pre-training process.}\\
\cline{2-4}          & S3VDC (2020) \cite{cao2020simple} &   \multicolumn{2}{l|}{ Improvement on four generic algorithmic.} \\
\cline{2-4}          & DSVAE (2021) \cite{9662055} & \multicolumn{2}{l|}{ Spherical latent embeddings.} \\
\cline{2-4}          & DVAE (2022) \cite{ma2022achieving} & \multicolumn{2}{l|}{Additional classifier to distinguish clusters.} \\

    \hline
    Net   & \multicolumn{1}{c|}{Methods} & With DAE & \multicolumn{1}{c|}{Characteristics} \\
    \hline
    \multirow{3}[14]{*}{GAN} & CatGAN (2015) \cite{springenberg2015unsupervised} & No    & Can be applied to both unsupervised and semi-supervised tasks. \\
\cline{2-4}          & DAGC (2017) \cite{harchaoui2017deep} & Yes   & Build an encoder to make the data representations easier to cluster. \\
\cline{2-4}          & DASC (2018) \cite{zhou2018deep} & Yes   & Subspace clustering. \\
\cline{2-4}          & ClusterGAN-SPL (2019) \cite{ghasedi2019balanced} & No    & No discrete latent variables and applies self-paced learning based on \cite{mukherjee2019clustergan}. \\
\cline{2-4}          & ClusterGAN (2019) \cite{mukherjee2019clustergan} & No    & Train a GAN with a clustering-specific loss. \\
\cline{2-4}          & ADEC (2020) \cite{mrabah2020adversarial} & Yes   & Reconstruction loss and adversarial loss are optimized in turn. \\
\cline{2-4}          & IMDGC (2022) \cite{9669125} & No   &  Integrates a hierarchical generative adversarial network and mutual information maximization. \\

    \hline
  
    \multicolumn{1}{|c|}{Net} & \multicolumn{1}{c|}{Methods} & \multicolumn{2}{c|}{Characteristics} \\
    \hline
    \multirow{5}[1]{*}{GNN}& DAEGC (2019) \cite{dilokthanakul2016deep} & \multicolumn{2}{l|}{ Perform graph clustering and learn graph embedding in a unified framework.} \\ 
\cline{2-4}          & AGC (2019) \cite{zhang2019attributed}                             &   \multicolumn{2}{l|}{ Attributed graph clustering.} \\
\cline{2-4}          & AGAE (2019) \cite{tao2019adversarial} &    \multicolumn{2}{l|}{Ensemble clustering.}\\
\cline{2-4}          & AGCHK (2020) \cite{zhu2020adaptive} &    \multicolumn{2}{l|}{Utilize heat kernel in attributed graphs.} \\
\cline{2-4}          & SDCN (2020) \cite{bo2020structural} &    \multicolumn{2}{l|}{  Integrate the structural information into deep clustering.} \\

\hline
    \end{tabular}%
  \label{The summaries of VAE-based and GAN-based methods in single-view clustering.}%
\end{table*}%

The autoencoder network \cite{bengio2013representation} is originally designed for unsupervised representation learning of data and can learn a highly non-linear mapping function. 
Using deep autoencoder (DAE) \cite{hinton2006reducing} is a common way to develop deep clustering methods.
DAE aims to learn a low-dimensional embedding feature space by minimizing the reconstruction loss of the network, which is defined as:
\begin{equation}
\label{AE loss}
L_{rec}=\min \frac{1}{n}\sum_{i=1}^n{\lVert x_i-\phi _r\left( \phi _e\left( x_i \right) \right) \rVert ^2}
\end{equation}
where ${\phi}_e(.)$ and ${\phi}_r(.)$ represent the encoder network and decoder network of autoencoder respectively. Using the encoder as a feature extractor, various clustering objective functions have been proposed. We summarize these deep autoencoder based clustering methods as \emph{DAE-based} deep clustering. In \emph{DAE-based} deep clustering methods, there are two main ways to get the labels. The first way embeds the data into low-dimensional features and then clusters the embedded features with traditional clustering methods such as the $k$-means algorithm\cite{kmeans}. The second way jointly optimizes the feature extractor and the clustering results. We refer to these two approaches as ``separate analysis'' and ``joint analysis'' respectively, and elaborate on them below. 

\par ``Separate analysis'' means that learning features and clustering data are performed separately.
In order to solve the problem that representations learned by ``separately analysis'' are not cluster-oriented due to its innate characteristics, Huang \emph{et al}. propose a deep embedding network for clustering (DEN) \cite{huang2014deep}, which imposes two constraints based on DAE objective: locality-preserving constraint and group sparsity constraint. Locality-preserving constraint urges the embedded features in the same cluster to be similar. Group sparsity constraint aims to diagonalize the affinity of representations. These two constraints improve the clustering performance while reduce the inner-cluster distance and expand inter-cluster distance. The objective of most clustering methods based on DAE are working on these two kinds of distance. So, in Table~\ref{The summaries of DAE-based and DNN-based methods in single-view clustering.}, we summarize these methods from the perspective of ``characteristics'', which shows the way to optimize the inner-cluster distance and inter-cluster distance.

Peng \emph{et al}. \cite{peng2016deep} propose a novel deep learning based framework in the field of Subspace clustering, namely, deep subspace clustering with sparsity prior (PARTY). PARTY enhances autoencoder by considering the relationship between different samples (i.e., structure prior) and solves the limitation of traditional subspace clustering methods. As far as we know, PARTY is the first deep learning based subspace clustering method, and it is the first work to introduce the global structure prior to the neural network for unsupervised learning. Different from PARTY, Ji \emph{et al}. \cite{ji2017deep} propose another deep subspace clustering networks (DSC-Nets) architecture to learn non-linear mapping and introduce a self-expressive layer to directly learn the affinity matrix. 

Density-based clustering \cite{ester1996density,rodriguez2014clustering} is another kind of popular clustering methods. Ren \emph{et al}. \cite{ren2020deep} propose deep density-based image clustering (DDIC) that uses DAE to learn the low-dimensional feature representations and then performs density-based clustering on the learned features. In particular, DDIC does not need to know the number of clusters in advance.

``Joint analysis'' aims at learning a representation that is more suitable for clustering which is different from separate analysis approaches that deep learning and clustering are carried out separately, the neural network does not have a clustering-oriented objective when learning the features of data. Most subsequent deep clustering researches combine clustering objectives with feature learning, which enables the neural network to learn features conducive to clustering from the potential distribution of data. In this survey, those methods are summarized as ``joint analysis''. 

Inspired by the idea of non-parametric algorithm t-SNE \cite{Maaten2009Learning}, Xie \emph{et al}. \cite{xie2016unsupervised} propose a joint framework to optimize feature learning and clustering objective, which is named deep embedded clustering (DEC). DEC firstly learns a mapping from the data space to a lower-dimensional feature space via $L_{rec}$ and then iteratively optimizes the clustering loss $KL(S\Vert R)$ (i.e., KL divergence).
Here, $S$ denotes the soft assignments of data that describes the similarity between the embedded data and each cluster centroid (centroids are initialized with $k$-means), and 
$R$ is the adjusted target distribution which has purer cluster assignments compared to $S$.

DEC is a representative method in deep clustering due to its joint learning framework and low computing complexity. Based on DEC, a number of variants have been proposed. For example, to guarantee local structure in the fine-tuning phase, improved deep embedded clustering with local structure preservation (IDEC) \cite{guo2017improved} is proposed to optimize the weighted clustering loss and the reconstruction loss of autoencoder jointly. Deep embedded clustering with data augmentation (DEC-DA) \cite{guo2018deep} applies the data augmentation strategy in DEC. Li \emph{et al}. \cite{li2018discriminatively} propose discriminatively boosted image clustering (DBC) to deal with image representation learning and image clustering. DBC has a similar pipeline as DEC but the learning procedure is self-paced \cite{kumar2010self}, where easiest instances are first selected and more complex samples are expanded progressively.

In DEC, the predicted clustering assignments are calculated by the Student’s $t$-distribution. Differently, Dizaji \emph{et al}. \cite{ghasedi2017deep} propose a deep embedded regularized clustering (DEPICT) with a novel clustering loss by stacking a softmax layer on the embedded layer of the convolutional autoencoder. What's more, the clustering loss of DEPICT is regularized by a prior for the frequency of cluster assignments and layer-wise features reconstruction loss function. 
Yang \emph{et al}. \cite{yang2017towards} directly take the objective of $k$-means as the clustering loss. The proposed model, named deep clustering network (DCN), is a joint dimensionality reduction and $k$-means clustering approach, in which dimensionality reduction is accomplished via learning a deep autoencoder. Shah \emph{et al}. \cite{shah2018deep} propose deep continuous clustering (DCC), an extension of robust continuous clustering \cite{shah2017robust} by integrating autoencoder into the paradigm. DCC performs clustering learning by jointly optimizing the defined data loss, pairwise loss, and reconstruction loss. In particular, it does not need prior knowledge of the number of clusters. 
Tzoreff \emph{et al} \cite{tzoreff2018deep} propose DDLSC (deep discriminative latent space for clustering) to optimize the deep autoencoder w.r.t. a discriminative pairwise loss function.

Deep manifold clustering (DMC) \cite{chen2017unsupervised} is the first method to apply deep learning in multi-manifold clustering \cite{souvenir2005manifold,elhamifar2011sparse}. In DMC, an autoencoder consists of stacked RBMs \cite{nair2010rectified} is trained to obtain the transformed representations. Both the reconstruction loss and clustering loss of DMC are different from previous methods. That is, the reconstruction of one sample and its local neighborhood are used to define the locality preserving objective. The penalty coefficient and the distance, measured by the Gaussian kernel between samples and cluster centers, are used to define the clustering-oriented objective. 


The recently proposed \emph{DAE-based} clustering algorithms also use the variants of deep autoencoder to learn better low-dimensional features and focus on improving the clustering performance by combining the ideas of traditional machine learning methods. 
For example,  deep spectral clustering using dual autoencoder network (DSCDAE) \cite{yang2019deep} and spectral clustering via ensemble deep autoencoder learning (SC-EDAE) \cite{affeldt2020spectral} aim to integrate spectral clustering into the carefully designed autoencoders for deep clustering. Zhang \emph{et al}. \cite{zhang2019neural} propose neural collaborative subspace clustering (NCSC) using two confidence maps, which are established on the features learned by autoencoder, as supervision information for subspace clustering. In ASPC-DA (adaptive self-paced deep clustering with data augmentation \cite{guo2020adaptive}), self-paced learning idea \cite{kumar2010self} and data augmentation technique are simultaneously incorporated. Its learning process is the same as DEC and consists of two stages, i.e., pre-training the autoencoder and fine-tuning the encoder.

In general, we notice that the network structure adopted is related to the type of data to be processed. For example, fully connected networks are generally used to extract one-dimensional data features, while convolutional neural networks are used to extract image features. Most of the above \emph{DAE-based} deep clustering methods can be implemented by both fully connected autoencoder and convolutional autoencoder, and thus they apply to various types of data to some extent. However, in the field of computer vision, there is a class of deep clustering methods that focus on image clustering. Those methods can date back to \cite{dundar2015convolutional} and are summarized as \emph{DNN-based} deep clustering because they generally use convolutional neural networks to perform image feature learning and semantic clustering.

\subsection{DNN-based}
\par
\begin{figure}[!t]
\centering
\includegraphics[scale=0.45]{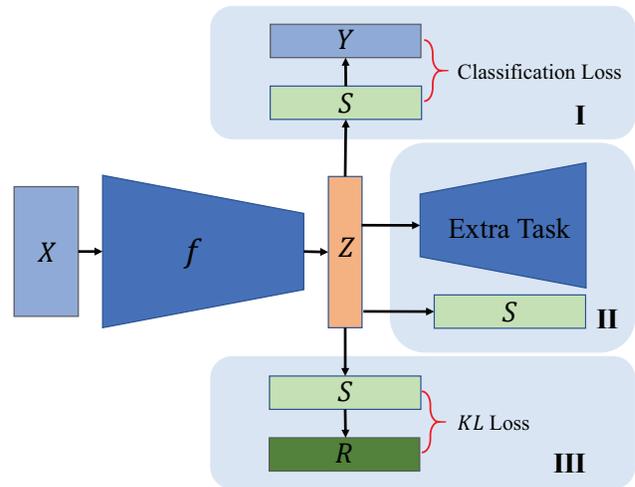}
\caption{The framework of \emph{DNN-based} learning (single-view clustering). $X$ is the data for clustering, $f$ is the feature extractor for $X$.  Part I describes the framework of supervised learning. $Y$ means the real labels and $S$ denotes the predicted results. With $Y$ and $S$, we can compute the classification loss for backpropagation.  Part II is the framework of methods with extra tasks. The extra tasks are used to train the nets for good embedding $Z$. Part III describes the process of the methods which need to fine-tune the cluster assignments. $S$ denotes the predicted results, $R$ is an adjustment of $S$. }
\label{fig:supervised}
\end{figure}

 This section introduces the \emph{DNN-based} clustering methods. Unlike \emph{DAE-based} clustering methods, \emph{DNN-based} methods have to design extra tasks to train the feature extractor. In this survey, we summarize \emph{DNN-based} deep clustering methods in Table \ref{The summaries of DAE-based and DNN-based methods in single-view clustering.} from two perspectives: ``clustering-oriented loss'' and ``characteristics''. ``clustering-oriented loss'' shows whether there is a loss function which explicitly narrows the inner-cluster distance or widens the inter-cluster distance. 
Fig.~\ref{fig:supervised} shows the framework of deep unsupervised learning based on a convolutional neural network.

When the DNN
training process begins, the randomly initialized feature extractor is unreliable.
So, deep clustering methods based on randomly initialized neural networks generally employ traditional clustering tricks such as hierarchical clustering \cite{johnson1967hierarchical} or focus on extra tasks such as instances generation. For instance, Yang \emph{et al}. \cite{yang2016joint} propose a joint unsupervised learning method named JULE, which applies agglomerative clustering magic to train the feature extractor. Specifically, JULE formulates the joint learning in a recurrent framework, where merging operations of agglomerative clustering are considered as a forward pass, and representation learning of DNN as a backward pass. Based on this assumption, JULE also applies a loss that shrinks the inner-cluster distance and expands the intra-cluster distance at the same time. In each epoch, JULE merges two clusters into one and computes the loss for the backward pass. 

Chang \emph{et al}. \cite{chang2017deep} propose deep adaptive image clustering (DAC) to tackle the combination of feature learning and clustering. In DAC, the clustering problem is reconstructed into binary pairwise classification problems that judge whether the pairwise images with estimated cosine similarities belong to the same cluster. Then it adaptively selects similar samples to train DNN in a supervised manner. DAC provides a novel perspective for deep clustering, but it only focuses on relationships between pairwise patterns. DDC (deep discriminative clustering analysis \cite{chang2019deep}) is a more robust and generalized version of DAC by introducing global and local constraints of relationships. ST-DAC ( spatial transformer - deep adaptive clustering \cite{souza2019improving}) applies a visual attention mechanism \cite{jaderberg2015spatial} to modify the structure of DAC. Haeusser \emph{et al}. \cite{haeusser2018associative} propose associative deep clustering (ADC), which contains a group of centroid variables with the same shape as image embeddings. With the intuition that centroid variables can carry over high-level information about the data structure in the iteration process, the authors introduce an objective function with multiple loss terms to simultaneously train those centroid variables and the DNN’s parameters along with a clustering mapping layer.

The above mentioned clustering methods estimate the cluster of an instance by passing it through the entire deep network, which tends to extract the global features of the instance \cite{krizhevsky2012imagenet}. Some clustering methods use mature classification network to initialize the feature extractor. For instance, DeepCluster \cite{caron2018deep} applies $k$-means on the output features of the deep model (like AlexNet \cite{krizhevsky2017imagenet} and VGG-16 \cite{simonyan2014very}) and uses the cluster assignments as ``pseudo-labels'' to optimize the parameters of the convolutional neural networks. Hsu \emph{et al}. \cite{hsu2018cnn} propose clustering CNN (CCNN)  which integrates mini-batch $k$-means with the model pretrained from the ImageNet dataset  \cite{deng2009imagenet}.

To improve the robustness of the model, more and more approaches make use of data augmentation for deep clustering \cite{guo2018deep,guo2020adaptive,huang2020deep}. 
For example, Huang \emph{et al}. \cite{huang2020deep} extend the idea of classical maximal margin clustering \cite{xu2004maximum,cortes1995support} to establish a novel deep semantic clustering method (named PartItion Confidence mAximisation - PICA). In PICA, three operations including color jitters, random rescale, and horizontal flip are adopted for data augmentation and perturbations.

Mutual information is also taken as a criterion to learn representations \cite{hu2017learning,hjelm2018learning} and becomes popular in recent clustering methods especially for image data. Various data augmentation techniques have been applied to generate transformed images that are used to mine their mutual information. For example, Ji \emph{et al}. \cite{ji2019invariant} propose invariant information clustering (IIC) for semantic clustering and image segmentation. In IIC, every image and its random transformation are treated as a sample pair. By maximizing mutual information between the clustering assignments of each pair, the proposed model can find semantically meaningful clusters and avoid degenerate solutions naturally. Instead of only using pairwise information, deep comprehensive correlation mining (DCCM) \cite{wu2019deep} is a novel image clustering framework, which uses pseudo-label loss as supervision information. Besides, the authors extend the instance level mutual information and present triplet mutual information loss to learn more discriminative features. Based on the currently fashionable contrastive learning \cite{chopra2005learning}, Zhong \emph{et al}. \cite{zhong2020deep} propose deep robust clustering (DRC), where two contrastive loss terms are introduced to decrease intra-class variance and increase inter-class variance. Mutual information and contrastive learning are related. In DRC, the authors summarize a framework that can turn maximize mutual information into minimizing contrastive loss.

In the field of image clustering on the semantic level, people think that the prediction of the original image should be consistent with that of the transformed image by data augmentation. So in the unsupervised learning context, data augmentation techniques not only are used to expand the training data but also can easily obtain supervised information. 
This is why data augmentation can be widely applied in many recent proposed image clustering methods. For example, Nina \emph{et al}. \cite{nina2019decoder} propose a decoder-free approach with data augmentation (called random triplet mining - RTM) for clustering and manifold learning. To learn a more robust encoder, the model consists of three encoders with shared weights and is a triplet network architecture conceptually. The first and the second encoders take similar images generated by data augmentation as positive pair, the second and the third encoders take a negative pair selected 
by RTM. Usually, the objective of triplet networks \cite{schroff2015facenet} is defined to make the features of the positive pair more similar and that of the negative pair more dissimilar. 

Although many existing deep clustering methods jointly learn the representations and clusters, such as JULE and DAC, there are specially designed representation learning methods \cite{doersch2015unsupervised,pathak2016context,zhang2016colorful,noroozi2016unsupervised,gidaris2018unsupervised} to learn the visual representations of images in a self-supervised manner. Those methods learn semantical representations by training deep networks to solve extra tasks. Such tasks can be predicting the patch context \cite{doersch2015unsupervised}, inpainting patches \cite{pathak2016context}, colorizing images \cite{zhang2016colorful}, solving jigsaw puzzles \cite{noroozi2016unsupervised}, and predicting rotations \cite{gidaris2018unsupervised}, etc. Recently, these self-supervised representation learning methods are adopted in image clustering. 
For example, MMDC (multi-modal deep clustering \cite{shiran2019multi}) leverages an auxiliary task of predicting rotations to enhance clustering performance. 
SCAN (semantic clustering by adopting nearest neighbors \cite{wvangansbeke2020scan}) first employs a self-supervised representation learning method to obtain semantically meaningful and high-level features. Then, it integrates the semantically meaningful nearest neighbors as prior information into a learnable clustering approach.

Since DEC \cite{xie2016unsupervised} and JULE \cite{yang2016joint} are proposed to jointly learn feature representations and cluster assignments by deep neural networks, many \emph{DAE-based} and \emph{DNN-based} deep clustering methods have been proposed and have made great progress in clustering tasks. However, the feature representations extracted in clustering methods are difficult to extend to other tasks, such as generating samples. 
The deep generative models have recently attracted a lot of attention because they can use neural networks to obtain data distributions so that samples can be generated (VAE \cite{kingma2013auto}, GAN \cite{goodfellow2014generative}, Pixel-RNN \cite{oord2016pixel}, InfoGAN \cite{chen2016infogan} and PPGN \cite{nguyen2017plug}). Specifically, GAN and VAE are the two most typical deep generative models. In recent years, researchers have applied them to various tasks, such as semi-supervised classification \cite{ehsan2017infinite,kingma2014semi,maaloe2016auxiliary,salimans2016improved}, clustering \cite{makhzani2015adversarial}, and image generation \cite{dosovitskiy2016generating,radford2015unsupervised}. In the next two subsections, we introduce the deep clustering algorithms based on the generated models: \emph{VAE-based} deep clustering and \emph{GAN-based} deep clustering.

\subsection{VAE-based}
Deep learning with nonparametric clustering (DNC) \cite{chen2015deeplearning} is a pioneer work in applying deep belief network to deep clustering. But in deep clustering based on the probabilistic graphical model, more research comes from the application of variational autoencoder (VAE), which combines variational inference and deep autoencoder together.

Most \emph{VAE-based} deep clustering algorithms aim at solving an optimization problem about evidence lower bound (ELBO, see the deduction details in \cite{kingma2013auto,hoffman2016elbo}), $p$ is the joint probability distribution, $q$ is the approximate probability distribution of $p(z|x)$, $x$ is the input data for clustering and $z$ the latent variable generated corresponding to $x$:
\begin{equation}
L_{\scriptscriptstyle ELBO}=\mathbb{E}_{q\left( z|x \right)}\left[ \log \frac{p\left( x,z \right)}{q\left( z|x \right)} \right] 
\end{equation}
The difference is that different algorithms have different generative models of latent variables or different regularizers. We list several \emph{VAE-based} deep clustering methods that have attracted much attention in recent years as below. For convenience, we omit the parameterized form of the probability distribution.
\par Traditional VAE generates a continuous latent vector $z$, $x$ is the vector of an original data sample. For the clustering task, the \emph{VAE-based} methods generate latent vector $(z,y)$, where $z$ is the latent vector representing the embedding and $y$ is the label. So the ELBO for optimization becomes:
\begin{equation}
L_{\scriptscriptstyle ELBO}=\mathbb{E}_{q\left( z,y|x \right)}\left[ \log \frac{p\left( x,z,y \right)}{q\left( z,y|x \right)} \right] 
\end{equation}
The first proposed unsupervised deep generative clustering framework is VaDE (variational deep embedding \cite{jiang2016variational}). VaDE models the data generative procedure with a GMM (Gaussian mixture model \cite{mclachlan2000finite}) combining a VAE. In this method, the cluster assignments and the latent variables are jointly considered in a Gaussian mixture prior rather than a single Gaussian prior.

Similar to VaDE, GMVAE (Gaussian mixture variational autoencoder \cite{dilokthanakul2016deep}) is another deep clustering method that combines VAE with GMM. Specifically, GMVAE considers the generative model $p(x,z,n,c) = p(x|z)p(z|n,c)p(n)p(c)$, where $c$ is uniformly distributed of $k$ categories and $n$ is normally distributed. $z$ is a continuous latent variable, whose distribution is a Gaussian mixture with means and variances of $c$ and $n$. Based on the mean-field theory \cite{beal2003variational}, GMVAE factors $q(z,n,c|x) = q(z|x)q(n|x)p(c|z,n)$ as posterior proxy. In the same way, those variational factors are parameterized with neural networks and the ELBO loss is optimized. 

On the basis of GMM and VAE, LTVAE (latent tree variational autoencoder \cite{li2018learning}) applies \emph{latent tree model} \cite{zhang2004hierarchical} to perform representation learning and structure learning for clustering. Differently, LTVAE has a variant of VAE with a superstructure of latent variables. The superstructure is a tree structure of discrete latent variables on top of the latent features. The connectivity structure among all variables is defined as a latent structure of the \emph{latent tree model} that is optimized via message passing \cite{koller2009probabilistic}. 

\textcolor{red}{
}
The success of some deep generative clustering methods depends on good initial pre-training. For example, in VaDE \cite{jiang2016variational}, pre-training is needed to initialize cluster centroids. In DGG \cite{yang2019deepgg}, pre-training is needed to initialize the graph embeddings. Although GMVAE \cite{dilokthanakul2016deep} learns the prior and posterior parameters jointly, the prior for each class is dependent on a random variable rather than the class itself, which seems counter-intuitive. 
Based on the ideas of GMVAE and VaDE, to solve their fallacies, Prasad \emph{et al}. \cite{prasad2020variational} propose a new model leveraging variational autoencoders for image clustering (VAEIC). Different from the methods mentioned above, the prior of VAEIC is deterministic, and the prior and posterior parameters are learned jointly without the need for a pre-training process. Instead of performing Bayesian classification as done in GMVAE and VaDE, VAEIC adopts more straight-forward inference and more principled latent space priors, leading to a simpler inference model $p(x,z,c)=p(x\vert z)p(z\vert c)p(c)$ and a simpler approximate posterior $q(z,c\vert x)=q(c\vert x)q(z\vert x,c)$. The cluster assignment is directly predicted by $q(c|z)$. What is more, the authors adopt data augmentation and design an image augmentation loss to make the model robust.

In addition to the \emph{VAE-based} deep clustering methods mentioned above, Figueroa \emph{et al}. \cite{figueroa2017simple} use the continuous Gumbel-Softmax distribution \cite{jang2016categorical,maddison2016concrete} to approximate the categorical distribution for clustering. Willetts \emph{et al}. \cite{willetts2019disentangling} extend variational ladder autoencoders \cite{zhao2017learning} and propose a disentangled clustering algorithm. Cao \emph{et al}. \cite{cao2020simple} propose a simple, scalable, and stable variational deep clustering algorithm, which introduces generic improvements for variational deep clustering. 

\subsection{GAN-based}

\begin{figure}[!t]
\centering
\includegraphics[scale=0.6]{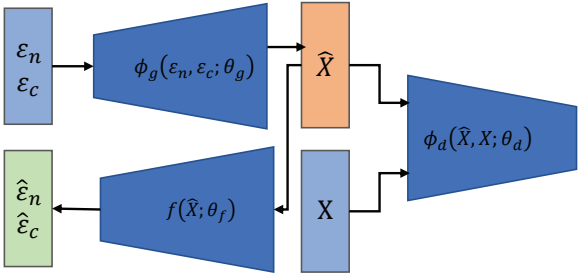}
\caption{The framework of \emph{GAN-based} learning. 
$\phi_g$ is the generator and $\phi_d$ is the discriminator, both $\varepsilon_n$ and $\varepsilon_c$ are inputs to the generator, $\varepsilon_n$ is the noise and $\varepsilon_n$ is the class information. $X$ is the data for clustering, $\hat{X}$ is the fake data which ``fools'' the discriminator, the function $f(·)$ operates on $\hat{X}$ to generate $\hat{\varepsilon_n}$ and $\hat{\varepsilon_c}$.
}
\label{fig:The framework of GAN-based learning.}
\end{figure}

In adversarial learning, standard generative adversarial networks (GANs) \cite{goodfellow2014generative} are defined as an adversarial game between two networks: generator $\phi_g$ and discriminator $\phi_d$. Specifically, the generator is optimized to generate fake data that “fools” the discriminator, and the discriminator is optimized to tell apart real from fake input data as shown in Fig. \ref{fig:The framework of GAN-based learning.}. 


GAN has already been widely applied in various fields of deep learning. Many deep clustering methods also adopt the idea of adversarial learning due to their strength in learning the latent distribution of data. We summarize the important \emph{GAN-based} deep clustering methods as follows.
Probabilistic clustering algorithms address many unlabeled data problems, such as regularized information maximization (RIM) \cite{krause2010discriminative}, or the related entropy minimization \cite{grandvalet2005semi}. The main idea of RIM is to train a discriminative classifier with unlabeled data. Unfortunately, these methods are prone to overfitting spurious correlations. Springenberg \emph{et al}. \cite{springenberg2015unsupervised} propose categorical generative adversarial networks (CatGAN) to address this weakness. To make the model more general, GAN is introduced to enhance the robustness of the classifier. In CatGAN, all real samples are assigned to one of the $k$ categories using the discriminator, while staying uncertain of clustering assignments for samples from the generative model rather than simply judging the false and true samples. In this way, the GAN framework is improved so that the discriminator can be used for multi-class classification. In particular, CatGAN can be applied to both unsupervised and semi-supervised tasks.

Interpretable representation learning in the latent space has been investigated in the seminal work of InfoGAN \cite{chen2016infogan}. Although InfoGAN does use discrete latent variables, it is not specifically designed for clustering. VAE \cite{kingma2013auto} can jointly train the inference network and autoencoder, which enables mapping from initial sample $X$ to latent space $Z$ that could potentially preserve cluster structure. Unfortunately, there is no such inference mechanism in GAN. 
To make use of their advantages, Mukherjee \emph{et al}. \cite{mukherjee2019clustergan} propose ClusterGAN as a new mechanism for clustering. ClusterGAN samples latent variables from a mixture of one-hot variables and continuous variables and establishes a reverse-mapping network to project data into a latent space.
It jointly trains a GAN along with the inverse-mapping network with a clustering-specific loss to achieve clustering. 

There is another \emph{GAN-based} deep clustering method \cite{ghasedi2019balanced} (we denote it as ClusterGAN-SPL) that has a similar network module with ClusterGAN. The main difference is that ClusterGAN-SPL does not set discrete latent variables but applies self-paced learning \cite{kumar2010self} to improve the robustness of the algorithm.

In some \emph{GAN-based} deep clustering methods (e.g., DAGC \cite{harchaoui2017deep}, DASC \cite{zhou2018deep}, AGAE \cite{tao2019adversarial} and ADEC \cite{mrabah2020adversarial}), generative adversarial network and deep autoencoder are both applied. For example, inspired by the adversarial autoencoders \cite{makhzani2015adversarial} and GAN \cite{goodfellow2014generative}, Harchaoui \emph{et al}. \cite{harchaoui2017deep} propose deep adversarial gaussian mixture autoencoder for clustering (DAGC). To make the data representations easier to cluster than in the initial space, it builds an autoencoder \cite{Vincent2010Stacked} consisting of an encoder and a decoder. In addition, an adversarial discriminator is added to continuously force the latent space to follow the Gaussian mixture prior \cite{mclachlan2000finite}. This framework improves the performance of clustering due to the introduction of adversarial learning. 

Most existing subspace clustering approaches ignore the inherent errors of clustering and rely on the self-expression of handcrafted representations. Therefore, their performance on real data with complex underlying subspaces is not satisfactory. Zhou \emph{et al}. \cite{zhou2018deep} propose deep adversarial subspace clustering (DASC) to alleviate this problem and apply adversarial learning into deep subspace clustering. DASC consists of a generator and a discriminator that learn from each other. The generator outputs subspace clustering results and consists of an autoencoder, a self-expression layer, and a sampling layer. The deep autoencoder and self-expression layer are used to convert the original input samples into better representations. In the pipeline, a new ``fake'' sample is generated by sampling from the estimated clusters and sent to the discriminator to evaluate the quality of the subspace cluster. 
\textcolor{red}{}

Many autoencoder based clustering methods use reconstruction for pretraining and let reconstruction loss be a regularizer in the clustering phase. Mrabah \emph{et al}. \cite{mrabah2020adversarial} point out that such a trade-off between clustering and reconstruction would lead to feature drift phenomena. Hence, the authors adopt adversarial training to address the problem and propose adversarial deep embedded clustering (ADEC). It first pretrains the autoencoder, where reconstruction loss is regularized by an adversarially constrained interpolation \cite{berthelot2018understanding}. Then, the cluster loss (similar to DEC \cite{xie2016unsupervised}), reconstruction loss, and adversarial loss are optimized in turn. ADEC can be viewed as a combination of deep embedded clustering and adversarial learning.

Besides the above-mentioned methods, there are a small number of deep clustering methods whose used networks are difficult to categorize. For example, IMSAT (information maximizing self-augmented training \cite{hu2017learning}) uses very simple networks to perform unsupervised discrete representation learning. SpectralNet \cite{shaham2018spectralnet} is a deep learning method to approximate spectral clustering, where unsupervised siamese networks \cite{hadsell2006dimensionality,shaham2018learning} are used to compute distances. In clustering tasks, it is a common phenomenon to adopt the appropriate neural network for different data formats. In this survey, we focus more on deep learning techniques that are reflected in the used systematic neural network structures.

\subsection{GNN-based}
Graph neural networks (GNNs) \cite{scarselli2008graph,duvenaud2015convolutional} allow end-to-end differentiable losses over data with arbitrary graph structure and have been applied to a wide range of applications. Many tasks in the real world can be described as a graph, such as social networks, protein structures, traffic networks, etc. With the suggestion of Banach’s fixed point theorem\cite{khamsi2011introduction}, GNN uses the following classic iterative scheme to compute the state. $F$ is a global transition function, the value of $H$ is the fixed point of $H=F(H,X)$ and is uniquely defined with the assumption that $F$ is a contraction map\cite{zhou2020graph}.
\begin{equation}
    H^{t+1}=F(H^{t},X)
\end{equation}
In the training process of GNN, many methods try to introduce attention and gating mechanism into a graph structure. Among these methods, graph convolutional network (GCN) \cite{kipf2016semi} which utilizes the convolution for information aggregation has gained remarkable achievement. $H$ is the node hidden feature matrix, $W$ is the learnable model parameters and $C$ is the feature matrix of a graph, the compact form of GCN is defined as:
\begin{equation}
    H=\widetilde{D}^{-\frac{1}{2}}\widetilde{Q}\widetilde{D}^{-\frac{1}{2}}CW
\end{equation}
 In the domain of unsupervised learning, there are also a variety of methods trying to use the powerful structure capturing capabilities of GNNs to improve the performance of clustering algorithms. We summarize the \emph{GNN-based} deep clustering methods as follows.

Tian \emph{et al.} propose DRGC (learning deep representations for graph clustering) \cite{tian2014learning} to replace traditional spectral clustering with sparse autoencoder and $k$-means algorithm. In DRGC, sparse autoencoder is adopted to learn non-linear graph representations that can approximate the input matrix through reconstruction and achieve the desired sparse properties. The last layer of the deep model outputs a sparse encoding and $k$-means serves as the final step on it to obtain the clustering results. To accelerate graph clustering, Shao \emph{et al.} propose deep linear coding for fast graph clustering (DLC) \cite{shao2015deep}. Unlike DRGC, DLC does not require eigen-decomposition and greatly saves running time on large-scale datasets, while still maintaining a low-rank approximation of the affinity graph.

The research on GNNs is closely related to graph embedding or network embedding\cite{cui2018survey,zhang2018network,cai2018comprehensive}, as GNNs can address the network embedding problem through a graph autoencoder framework\cite{wu2020comprehensive}.
The purpose of graph embedding \cite{yan2006graph} is to find low-dimensional features that maintain similarity between the vertex pairs in a sample similarity graph. If two samples are connected in the graph, their latent features will be close. Thus, they should also have similar cluster assignments. Based on this motivation, Yang \emph{et al}. \cite{yang2019deepgg} propose deep clustering via a Gaussian mixture variational autoencoder with graph embedding (DGG). Like VaDE \cite{jiang2016variational}, the generative model of DGG is $p(x,z,c) = p(x\vert z)p(z\vert c)p(c)$. The prior distributions of $z$ and $c$ are set as a Gaussian mixture distribution and a categorical distribution, respectively. The learning problem of GMM-based VAE is usually solved by maximizing the evidence lower bound (ELBO) of the log-likelihood function with \emph{reparameterization trick}. To achieve graph embedding, the authors add a graph embedding constraint to the original optimization problem, which exists not only on the features but also on the clustering assignments. Specifically, the similarity between data points is measured with a trained Siamese network \cite{hadsell2006dimensionality}.

\begin{figure*}
    \centering
    \includegraphics[width=17cm]{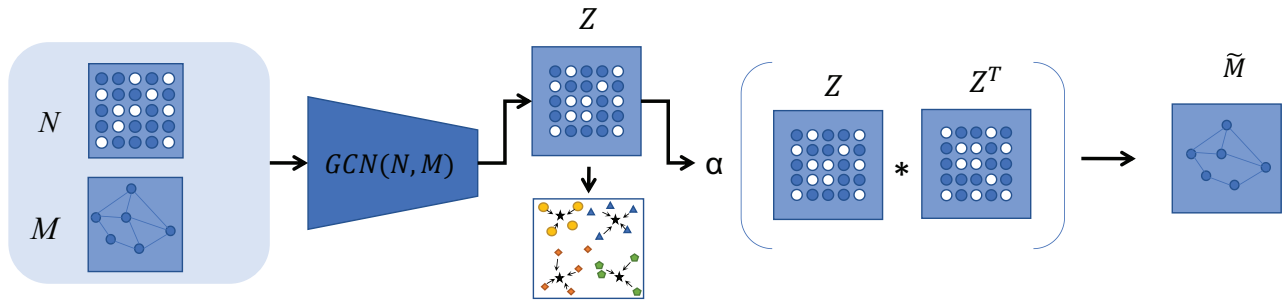}
    \caption{The data stream framework of graph autoencoder applicated in clustering. $GCN(N,M)$ is a graph autoencoder, $GCN(·)$ is used to represent a graph convolutional neural network, graph autoencoder consists of two layers of graph convolutional neural networks. Both node attributes $N$ and graph structure $M$ are utilized as inputs to this encoder. $Z$ is a matrix of node embedding vectors. $\alpha$ is an activation function, $\widetilde{M}$ is the prediction of graph adjacency matrix $M$.}
    \label{fig:GAE}
    \end{figure*}

Autoencoder also works on graphs as an effective embedding method. In AGAE (adversarial graph autoEncoders) \cite{tao2019adversarial}, the authors apply ensemble clustering \cite{Strehl2002cluster,fred2005combining} in the deep graph embedding process and develop an adversarial regularizer to guide the training of the autoencoder and discriminator.
Recent studies have mostly focused on the methods which are two-step approaches. The drawback is that the learned embedding may not be the best fit for the clustering task. To address this, Wang \emph{et al.} propose a unified approach named deep attentional embedded graph clustering (DAEGC) \cite{wang2019attributed}. DAEGC develops a graph attention-based autoencoder to effectively integrate both structure and content information, thereby achieving better clustering performance. The data stream framework of graph autoencoder applicated in clustering in Fig. \ref{fig:GAE}.

As one of the most successful feature extractors for deep learning, CNNs are mainly limited by Euclidean data. 
GCNs have proved that graph convolution is effective in deep clustering, e.g., Zhang \emph{et al.} propose an adaptive graph convolution (AGC)\cite{zhang2019attributed} method for attributed graph clustering. AGC exploits high-order graph convolution to capture global cluster structure and adaptively selects the appropriate order for different graphs. Nevertheless, AGC  might not determine the appropriate neighborhood that reflects the relevant information of connected nodes represented in graph structures. Based on AGC, Zhu \emph{et al.} exploit heat kernel to enhance the performance of graph convolution and propose AGCHK (AGC using heat kernel) \cite{zhu2020adaptive}, which could make the low-pass performance of the graph filter better. 

In summary, we can realize the importance of the structure of data. Motivated by the great success of GNNs in encoding the graph structure, Bo \emph{et al.} propose a structural deep clustering network (SDCN) \cite{bo2020structural}. By stacking multiple layers of GNN, SDCN is able to capture the high-order structural information. At the same time, benefiting from the self-supervision of AE and GNN, the multi-layer GNN does not exhibit the so-called over-smooth phenomenon. SDCN is the first work to apply structural information into deep clustering explicitly.

\section{Semi-supervised Deep Clustering}

\begin{table}[!h]
  \centering
  \caption{Semi-supervised deep clustering methods.}
    \begin{tabular}{|r|l|}
    \hline
    \multicolumn{1}{|c|}{Methods} & \multicolumn{1}{c|}{Characteristics}\\
    \hline
     SDEC (2019) \cite{ren2019semi} & Based on DEC \cite{xie2016unsupervised}. \\    \hline
SSLDEC (2019) \cite{enguehard2019semi} & Based on DEC \cite{xie2016unsupervised}. \\    \hline
DECC (2019) \cite{zhang2019framework} & Based on DEC \cite{xie2016unsupervised}. \\
    \hline
    SSCNN (2020) \cite{shukla2020semi} & Combine $k$-means loss and pairwise divergence. \\
    \hline
    \end{tabular}%
  \label{deep semi-supervised clustering}%
\end{table}%

\par Traditional semi-supervised learning can be divided into three categories, i.e., semi-supervised classification \cite{chapelle2005semi,huang2008semi}, semi-supervised dimension reduction \cite{xu2010discriminative,huang2012semi}, and
semi-supervised clustering \cite{basu2002semi,grira2004unsupervised,ren2018semi}. 
Commonly, the constraint of unsupervised data is marked as ``must-link'' and ``cannot-link''. Samples with the ``must-link'' constraint belong to the same cluster, while samples with the ``cannot-link'' constraint belong to different clusters. Most semi-supervised clustering objectives are the combination of unsupervised clustering loss and constraint loss.

Semi-supervised deep clustering has not been explored well. Here we introduce several representative works. These works use different ways to combine the relationship constraints and the neural networks to obtain better clustering performance. We summarize these methods in Table~\ref{deep semi-supervised clustering}.

Semi-supervised deep embedded clustering (SDEC) \cite{ren2019semi} is based on DEC \cite{xie2016unsupervised} and incorporates pairwise constraints in the feature learning process. Its loss function is defined as:
\begin{equation}
\begin{gathered}
  Loss = KL(S\parallel {R}){\text{ + }}\lambda \frac{{\text{1}}}{n}\sum\limits_{i = 1}^n {\sum\limits_{k = 1}^n {{a_{ij}}\parallel {z_i} - {z_j}{\parallel ^2}} } , \hfill \\
\end{gathered}
\end{equation}
where $\lambda$ is a trade-off parameter.
$a_{ij}$ = 1 if $x_i$ and $x_j$ are assigned to the same cluster, $a_{ij}$ = -1 if $x_i$ and $x_j$ satisfy cannot-link constraints, $a_{ij}$ = 0 otherwise. 
As the loss function shows, it is formed by two parts. The first part is KL divergence loss which has been explained in Section \ref{sec:DAE-based}. The second part is semi-supervised loss denotes the consistency between the embedded feature $\{ {z_i}\} _{i = 1}^n$ and parameter $a_{ij}$. Intuitively, if  $a_{ij}=1$, to minimize the loss function, ${\parallel}{z_i} - {z_j}{\parallel ^2}$ should be small. In contrast, if  $a_{ij}=-1$, to minimize the loss, ${\parallel}{z_i} - {z_j}{\parallel ^2}$ should be large, which means $z_i$ is apart from $z_j$ in the latent space $Z$.

\par Like SDEC, most semi-supervised deep clustering (DC) methods are based on unsupervised DC methods. It is straightforward to expand an unsupervised DC method to a semi-supervised DC one through adding the semi-supervised loss. Compared with unsupervised deep clustering methods, the extra semi-supervised information of data can help the neural network to extract features more suitable for clustering.  There are also some works focusing on extending the existing semi-supervised clustering method to a deep learning version. 
For example, the feature extraction process of both SSLDEC (semi-supervised learning with deep embedded clustering for image classification and segmentation) \cite{enguehard2019semi} and DECC (deep constrained clustering) \cite{zhang2019framework}  are based on DEC. Their training process is similar to semi-supervised $k$-means \cite{basu2002semi} which learns feature representations by alternatively using labeled and unlabeled data samples. During the training process, the algorithms use labeled samples to keep the model consistent and choose a high degree of confidence unlabeled samples as newly labeled samples to tune the network. Semi-supervised clustering with neural networks \cite{shukla2020semi} combines a $k$-means loss and pairwise divergence to simultaneously learn the cluster centers as well as semantically meaningful feature representations. 

\section{Deep Multi-view Clustering}
\par 
The above-mentioned deep clustering methods can only deal with single-view data. In practical clustering tasks, the input data usually have multiple views. For example, the report of the same topic can be expressed with different languages; the same dog can be captured from different angles by the cameras; the same word can be written by people with different writing styles. Multi-view clustering (MVC) methods \cite{chaudhuri2009multi,kumar2011co,li2015large,cao2015diversity,nie2017self,zhang2017latent,zhang2018binary,zhao2017multi,brbic2018multi,ren2019selfMVC,xu2015multi} are proposed to make use of the complementary information among multiple views to improve clustering performance. 

In recent years, the application of deep learning in multi-view clustering is a hot topic \cite{xu2021deep,fan2020one2multi,tang2018deep,li2019reciprocal,zhu2019multi}. Those deep multi-view clustering algorithms focus on solving the clustering problems with different forms of input data. Since the network structures used in most of these methods are autoencoders, we divided them into three categories based on the adopted clustering theoretical basis: \emph{
DEC-based}, \emph{subspace clustering-based}, and \emph{GNN-based}. They are summarized in Table~\ref{tab:dmvc}.

\begin{table*}[!t]
\caption{The summaries of deep multi-view clustering methods.}
\label{tab:dmvc}
\centering
\begin{tabular}{|c|r|l|l|}
\hline
Networks  & \multicolumn{1}{c|}{Methods}  & \multicolumn{1}{c|}{Characteristics} \\
\hline

DAE + GAN & DAMC (2019) \cite{li2019deep}        & Capture the data distribution ulteriorly by adversarial training.  \\
\cline{1-3}
VAE       & DMVCVAE (2020) \cite{yin2020shared} & Learn a shared latent representation under the VAE framework.                                    \\
\cline{1-3}
DAE       & DEMVC (2021) \cite{xu2021deep}  &  Through 
collaborative training, each view can guide all views.                                    \\
\cline{1-3}   
DAE       & DMVSSC (2018) \cite{tang2018deep} & Extract multi-view deep features by CCA-guided convolutional auto-encoders.                                         \\
\cline{1-3}
DAE       & RMSL (2019) \cite{li2019reciprocal} & Recover the underlying low-dimensional subspaces in which the high dimensional data lie.                                         \\
\cline{1-3}
DAE       & MVDSCN (2019) \cite{zhu2019multi} & Combine convolutional auto-encoder and self-representation together.                                         \\
\cline{1-3}
VAE & Multi-VAE (2021) \cite{xu2021multi} & Learn disentangle and explainable representations.\\
\cline{1-3}

DAE       & CMHHC (2022) \cite{lin2022contrastive} & Employ multiple autoencoders and hyperbolic hierarchical clustering.                       \\
\cline{1-3}
DAE       & MFLVC (2022) \cite{xu2022multi} & Utilize contrastive clustering to learn the common semantics across all views.                    \\
\cline{1-3}
DAE       & DIMVC (2022) \cite{xu2022deep} & Imputation-free and fusion-free incomplete multi-view clustering. \\
\cline{1-3}

\hline

\hline
GCN       & Multi-GCN (2019) \cite{khan2019multi} & Incorporates nonredundant information from multiple views.
            \\
            \cline{1-3}
GCN       & MAGCN (2020) \cite{cheng2020multi} &Dual encoders for reconstructing and integrating.
            \\
            \cline{1-3}
\hline
GAE       & O2MAC (2020) \cite{fan2020one2multi} &Partition the graph into several nonoverlapping clusters.
            \\
            \cline{1-3}
GAE       & CMGEC (2021) \cite{wang2021consistent} &Multiple graph autoencoder.
            \\
            \cline{1-3}
GAE       & DMVCJ (2022) \cite{huang2022deep} &Weighting strategy to alleviate the noisy issue.
            \\
            \cline{1-3}
\hline

\end{tabular}
\end{table*}


\subsection{DEC-based}
As mentioned previously, DEC (deep embedded clustering) \cite{xie2016unsupervised} uses autoencoder to learn the low-dimensional embedded feature representation and then minimizes the KL divergence of student's $t$-distribution and auxiliary target distribution of feature representations to achieve clustering. Improved DEC (IDEC) \cite{guo2017improved} emphasizes data structure preservation and adds the term of the reconstruction loss for the lower-dimensional feature representation when processing fine-tuning tasks. Some deep multi-view clustering methods also adopt this deep learning pipeline.

Traditional MVC methods mostly use linear and shallow embedding to learn the latent structure of multi-view data. These methods cannot fully utilize the non-linear property of data, which is vital to reveal a complex clustering structure. Based on adversarial learning and deep autoencoder, Li \emph{et al}. \cite{li2019deep} propose deep adversarial multi-view clustering (DAMC) to learn the intrinsic structure embedded in multi-view data. Specifically, DAMC consists of a multi-view encoder $E$, a multi-view generator (decoder) ${\phi}_g$, $V$ discriminators $D_1,...,D_V$ ($V$ denotes the number of views), and a deep embedded clustering layer. The multi-view encoder outputs low-dimensional embedded features for each view. For each embedded feature, the multi-view generator generates a corresponding reconstruction sample. The discriminator is used to identify the generated sample from the real sample and output feedback. The total loss function of DAMC is defined as:
\begin{equation}
    Loss = \min_{E,G} \max_{D_1,...,D_V} L_r + \alpha L_{c} + \beta L_{GAN},
\end{equation}
\noindent where $L_{c}$ comes from DEC \cite{xie2016unsupervised} and represents the clustering loss, $L_r$ and $L_{GAN}$ represent the reconstruction loss and GAN loss respectively, $\alpha$ and $\beta$ are hyperparameters. Compared with traditional MVC algorithms, DAMC can reveal the non-linear property of multi-view data and achieve better clustering performance.

Xu \emph{et al}. \cite{xu2021deep} propose a novel collaborative training framework for deep embedded multi-view clustering (DEMVC). Specifically, DEMVC defines a switched shared auxiliary target distribution and fuses it into the overall clustering loss. Its main idea is that by sharing optimization objectives, each view, in turn, guides all views to learn the low-dimensional embedded features that are conducive to clustering. At the same time, optimizing reconstruction loss makes the model retain discrepancies among multiple views. Experiments show that DEMVC can mine the correct information contained in multiple views to correct other views, which is helpful to improve the clustering accuracy. Existing methods tend to fuse multiple views’ representations, Xu \emph{et al}. \cite{xu2021multi} present a novel \emph{VAE-based} multi-view clustering framework (Multi-VAE) by learning disentangled visual representations.

Lin \emph{et al}. \cite{lin2022contrastive} propose a contrastive multi-view hyperbolic hierarchical clustering (CMHHC). It consists of three components, multi-view alignment learning, aligned feature similarity learning, and continuous hyperbolic hierarchical clustering. Through capturing the invariance information across views and learn the meaningful metric property for similarity-based continuous hierarchical clustering. CMHHC is capable of clustering multiview data at diverse levels of granularity. 

Xu \emph{et al}. \cite{xu2022multi} propose a framework of multi-level feature learning for contrastive multi-view clustering (MFLVC), which combines multi-view clustering with contrastive learning to improve clustering effectiveness. MFLVC can learn different levels of features and reduce the adverse influence of view-private information. Xu \emph{et al}. \cite{xu2022deep} also explore incomplete multi-view clustering, through mining the complementary information in the high-dimensional feature space via a nonlinear mapping of multiple views, the proposed method DIMVC can handle the incomplete data primely.

\subsection{Subspace clustering-based}
Subspace clustering\cite{vidal2011subspace} is another popular clustering method, which holds the assumption that data points of different clusters are drawn from multiple subspaces. Subspace clustering typically firstly estimates the affinity of each pair of data points to form an affinity matrix, and then applies spectral clustering \cite{ng2002spectral} or a normalized cut \cite{shi2000normalized} on the affinity matrix to obtain clustering results. Some subspace clustering methods based on self-expression \cite{elhamifar2009sparse} have been proposed. The main idea of self-expression is that each point can be expressed with a linear combination $C$ of the data points $X$ themselves. The general objective is:
\begin{equation}
    Loss=L_r + \alpha R(C) = \Vert X-XC \Vert + \alpha R(C),
\end{equation}
where $\Vert X-XC \Vert$ is the reconstruction loss and $R(C)$ is the regularization term for subspace representation $C$. In recent years, a lot of works \cite{nie2011unsupervised,lu2012robust,elhamifar2013sparse,liu2012robust,feng2014robust,peng2015robust,zhang2015low} generate a good affinity matrix and achieve better results by using the self-expression methodology.

There are also multi-view clustering methods \cite{cao2015diversity,zhang2017latent,brbic2018multi} which are based on subspace learning. They construct the affinity matrix with shallow features and lack of interaction across different views, thus resulting in insufficient use of complementary information included in multi-view datasets. To address this,   researchers focus more on multi-view subspace clustering methods based on deep learning recently. 

Exploring the consistency and complementarity of multiple views is a long-standing important research topic of multi-view clustering \cite{xu2013survey}. Tang \emph{et al}. \cite{tang2018deep} propose the deep multi-view sparse subspace clustering (DMVSSC), which consists of a canonical correlation analysis (CCA) \cite{anderson1962introduction,andrew2013deep,wang2015deep} based self-expressive module and convolutional autoencoders (CAEs). The CCA-based self-expressive module is designed to extract and integrate deep common latent features to explore the complementary information of multi-view data. A two-stage optimization strategy is used in DMVSSC. 
Firstly, it only trains CAEs of each view to obtain suitable initial values for parameters. Secondly, it fine-tunes all the CAEs and CCA-based self-expressive modules to perform multi-view clustering.

Unlike CCA-based deep MVC methods (e.g., DMVSSC \cite{tang2018deep}) which project multiple views into a common low-dimensional space, Li \emph{et al}. \cite{li2019reciprocal} present a novel algorithm named reciprocal multi-layer subspace learning (RMSL). RMSL contains two main parts: HSRL (hierarchical self-representative layers) and BEN (backward encoding networks). The self-representative layers (SRL) contains the view-specific SRL which maps view-specific features into view-specific subspace representations, and the common SRL which further reveals the subspace structure between the common latent representation and view-specific representations. BEN implicitly optimizes the subspaces of all views to explore consistent and complementary structural information to get a common latent representation. 
 
Many multi-view subspace clustering methods first extract hand-crafted features from multiple views and then learn the affinity matrix jointly for clustering. This independent feature extraction stage may lead to the multi-view relations in data being ignored. To alleviate this problem, Zhu \emph{et al}. \cite{zhu2019multi} propose a novel multi-view deep subspace clustering network (MVDSCN) which consists of diversity net (Dnet) and universality net (Unet). Dnet is used to learn view-specific self-representation matrices and Unet is used to learn a common self-representation matrix for multiple views. The loss function is made up of the reconstruction loss of autoencoders, the self-representation loss of subspace clustering, and multiple well-designed regularization items. 

\subsection{GNN-based}
In the real world, graph data are far more complex. For example, we can use text, images and links to describe the same web page, or we can ask people with different styles to write the same number. Obviously, traditional single-view clustering methods are unable to meet the needs of such application scenarios. That is, one usually needs to employ a multi-view graph\cite{qu2017attention}, rather than a single-view graph, to better represent the real graph data.
Since GCN has made considerable achievements in processing graph-structured data, Muhammad \emph{et al}. develop a graph-based convolutional network (Multi-GCN) \cite{khan2019multi} for multi-view data. Multi-GCN focuses attention on integrating subspace learning approaches with recent innovations in graph convolutional networks, and proposes an efficient method for adapting graph-based semi-supervised learning (GSSL) to multiview contexts. 

Most GNNs can effectively process single-view graph data, but they can not be directly applied to multi-view graph data. Cheng \emph{et al}. propose multi-view attribute graph convolution networks for clustering (MAGCN) \cite{cheng2020multi} to handle graph-structured data with multi-view attributes. The main innovative method of MAGCN is designed with two-pathway encoders. The first pathway develops multiview attribute graph attention networks to capture the graph embedding features of multi-view graph data. Another pathway develops consistent embedding encoders to capture the geometric relationship and the consistency of probability distribution among different views. 

Fan \emph{et al}. \cite{fan2020one2multi} attempt to employ deep embedded learning for multi-view graph clustering. The proposed model is named One2Multi graph autoencoder for multi-view graph clustering (O2MAC), which utilizes graph convolutional encoder of one view and decoders of multiple views to encode the multi-view attributed graphs to a low-dimensional feature space. Both the clustering loss and reconstruction loss of O2MAC are similar to other deep embedded clustering methods in form. What's special is that graph convolutional network \cite{kipf2016semi} is designed to deal with graph clustering tasks \cite{schaeffer2007graph}. 
Huang \emph{et al.} \cite{huang2022deep} propose DMVCJ (deep embedded multi-view clustering via jointly learning latent representations and graphs). By introducing a self-supervised GCN module, DMVCJ jointly learns both latent graph structures and feature representations. 


The graph in most existing GCN-based multi-view clustering methods is fixed, which makes the clustering performance heavily dependent on the predefined graph. A noisy graph with unreliable connections can result in ineffective convolution with wrong neighbors on the graph \cite{yun2019graph}, which may worsen the performance. To alleviate this issue, Wang \emph{et al}. propose a consistent multiple graph embedding clustering framework (CMGEC) \cite{wang2021consistent}, which is mainly composed of multiple graph autoencoder (M-GAE), multi-view mutual information maximization module (MMIM), and graph fusion network (GFN). CMGEC develops a multigraph attention fusion encoder to adaptively learn a common representation from multiple views, and thereby CMGEC can deal with three types of multi-view data, including multi-view data without a graph, multi-view data with a common graph, and single-view data with multiple graphs.

According to our research, deep multi-view clustering algorithms have not been explored well. Other than the above-mentioned three categories, Yin \emph{et al}. \cite{yin2020shared} propose a \emph{VAE-based} deep MVC method (deep multi-view clustering via variational autoencoders, DMVCVAE). DMVCVAE learns a shared generative latent representation that obeys a mixture of Gaussian distributions and thus can be regarded as the promotion of VaDE \cite{jiang2016variational} in multi-view clustering. There are also some application researches based on deep multi-view clustering. For example, Perkins \emph{et al}. \cite{perkins2019dialog} introduce the dialog intent induction task and present a novel deep multi-view clustering approach to tackle the problem. Abavisani \emph{et al}. \cite{abavisani2018deep} and Hu \emph{et al}. \cite{hu2019deep} study multi-modal clustering, which is also related to multi-view clustering. Taking advantage of both deep clustering  and multi-view learning will be an interesting future research direction of deep multi-view clustering.

\section{Deep Clustering with Transfer Learning}
\par Transfer learning has emerged as a new learning framework to address the problem that the training and testing data are drawn from different feature spaces or distributions \cite{pan2010a}.
For complex data such as high-resolution real pictures of noisy videos, traditional clustering methods even deep clustering methods can not work very well because of the high dimensionality of the feature space and no uniform criterion to guarantee the clustering process. Transfer learning provides new solutions to these problems through transferring the information from source domain that has additional information to guide the clustering process of the target domain. In the early phase, the ideas of deep domain adaption are simple and clear, such as DRCN (deep reconstruction-classification networks) \cite{ghifary2016deep} uses classification loss for the source domain and reconstruction loss for target domain. The two domains share the same feature extractor. With the development of DNN, we now have more advanced ways to transfer the knowledge.

In this section, we introduce some transfer learning work about clustering which are separated into two parts. The first part is ``\emph{DNN-based}'', and the second part is ``\emph{GAN-based}''. 
\subsection{DNN-based}

\par \emph{DNN-based} UDA methods generally aim at projecting the source and target domains into the same feature space, in which the classifier trained with source embedding and labels can be applied to the target domain. 



In 2014, through a summary of the network training processes, Yosinski \emph{et al}. \cite{yosinski2014transferable} find that many deep neural networks trained on natural images exhibit a phenomenon in common: the features learned in the first several layers appear not to be specific to a particular dataset or task and applicable to many other datasets or tasks. Features must eventually transition from general to specific by the last layers of the network. 
Thus, we can use a mature network  (e.g., AlexNet\cite{krizhevsky2017imagenet}, GoogleNet\cite{szegedy2015going}) which can provide credible parameters as the initialization for a specific task. This trick has been frequently used in feature extracted networks. 


Domain adaptive neural network (DaNN) \cite{ghifary2014domain} first used maximum mean discrepancy (MMD)\cite{gretton2012kernel} with DNN. 


\begin{figure}[!t]
\centering
\includegraphics[scale=0.45]{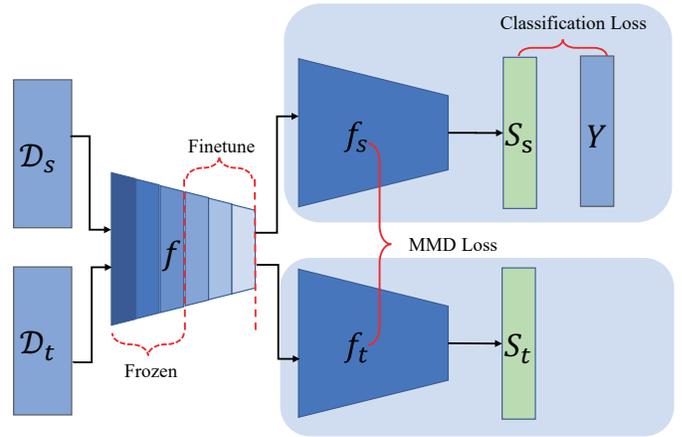}
\caption{
The data stream framework of deep adaption network (DAN). $\mathcal{D}_s$ is the source domain. $\mathcal{D}_t$ is the target domain. $f$ is the shared encoder of both domains, which can be initialized with existing network. The first layers of $f$ are frozen, the last layers of $f$ can be finetuned in the training process. $f_s$ is the encoder of $\mathcal{D}_s$. $f_t$ is the encoder of $\mathcal{D}_t$. $S_s$ are the predicted label vector of $\mathcal{D}_s$. $Y$ are the real labels of $\mathcal{D}_s$.  $S_t$ are the predicted results of $\mathcal{D}_t$. 
}
\label{fig:DAN}
\end{figure}

\par Many domain-discrepancy-based methods adopt similar techniques with DaNN. Deep adaption networks (DAN) \cite{long2015learning} use multiple kernel variants of MMD (MK-MMD) as its domain adaption function. As shown in Fig.~\ref{fig:DAN}, the net of DAN minimizes the distance at the last feature-specific layers and then the features from source-net and target-net would be projected into the same space. After DAN, more and more methods based on MMD are proposed. The main optimization way is to choose different versions of MMD, such as joint adaption network (JAN) \cite{long2017deep} and weighted deep adaption network (WDAN) \cite{yan2017mind}. JAN 
maximizes joint MMD to make the distributions of both source and target domains more distinguishable. 
WDAN is proposed to solve the question about imbalanced data distribution by introducing an auxiliary weight for each class in the source domain. RTN (unsupervised domain adaptation with residual transfer networks) \cite{long2016unsupervised} uses residual networks and MMD for UDA task. 
\par Some discrepancy-based methods do not use MMD. Domain adaptive hash (DAH)\cite{venkateswara2017deep} uses supervised hash loss and unsupervised entropy loss to align the target hash values to their corresponding source categories. Sliced wasserstein discrepancy (SWD) \cite{lee2019sliced} adopts the novel SWD to capture the dissimilarity of probability. Correlation alignment (CORAL) \cite{sun2017correlation} minimizes domain shift by aligning the second-order statistics of source and target distributions. Higher-order moment matching (HoMM) \cite{chen2020homm} shows that the first-order HoMM is equivalent to MMD and the second-order HoMM is equivalent to CORAL. 
Contrastive adaptation network (CAN) \cite{kang2019contrastive} proposes contrastive domain discrepancy (CDD) to minimize the intra-class discrepancy and maximize the inter-class margin. Besides, several new measurements are proposed for the source and target domain \cite{hu2020unsupervised,li2020enhanced,xu2020reliable}.
Analysis of representations for domain adaptation \cite{ben2006analysis} contributes a lot in the domain adaption distance field. 
Some works try to improve the performance of UDA in other directions, such as unsupervised domain adaptation via structured prediction based selective pseudo-labeling tries to learn a domain invariant subspace by supervised locality preserving projection (SLPP) using both labeled source data
and pseudo-labeled target data.

\par The tricks used in deep clustering have also been used in UDA methods. For example, structurally regularized deep clustering (SRDC) \cite{tang2020unsupervised} implements the structural source regularization via a simple strategy of joint network training. It first minimizes the KL divergence between the auxiliary distribution (that is the same with the auxiliary distribution of DEC \cite{xie2016unsupervised}) and the predictive label distribution. Then, it replaces the auxiliary distribution with that of ground-truth labels of source data. 
Wang \emph{et al}. \cite{wang2020unsupervised} propose a UDA method that uses novel selective pseudo-labeling strategy and learns domain invariant subspace by supervised locality preserving projection (SLPP) \cite{he2003locality} using both labeled source data and pseudo-labeled target data. Zhou \emph{et al}. \cite{zhou2021domain} apply ensemble learning in the training process. Prabhu \emph{et al}. \cite{prabhu2021sentry} apply entropy optimization in target domain.

\begin{table*}[htbp]
  \centering
  \caption{The summaries of \emph{DNN-} and \emph{GAN-based} methods in deep clustering with transfer learning.}
    \begin{tabular}{|c|r|l|}
    \hline
    Net   & \multicolumn{1}{c|}{Methods} & \multicolumn{1}{c|}{Characteristics} \\
    \hline
    \multirow{20}[9]{*}{DNN} & DaNN (2014) \cite{ghifary2014domain} & MMD and the same feature extracter. \\
\cline{2-3}          & DAN (2015) \cite{long2015learning} & Multi-kernel MMD. Different feature extracters. \\
\cline{2-3}          & DRCN (2016) \cite{ghifary2016deep} & Classification of source  and reconstruction of target. \\
\cline{2-3}          & RTN (2016) \cite{long2016unsupervised} & Residual networks and MMD. \\
\cline{2-3}          & DAH (2017) \cite{venkateswara2017deep} & Supervised hash loss and unsupervised  entropy loss. \\
\cline{2-3}          & WDAN (2017) \cite{yan2017mind} & Imbalanced data distribution. \\
\cline{2-3}          & JAN (2017) \cite{long2017deep} & Joint  MMD. \\
\cline{2-3}          & CORAL (2017) \cite{sun2017correlation} & Minimize domain shift by aligning the second-order statistics of source and target distributions. \\
\cline{2-3}          & SWD (2019) \cite{lee2019sliced} & Sliced Wasserstein discrepancy. \\
\cline{2-3}          & CAN (2019) \cite{kang2019contrastive} & Contrastive Domain Discrepancy. \\
\cline{2-3}          & SRDC (2020) \cite{tang2020unsupervised} & KL divergence and auxiliary distribution (the same with DEC \cite{xie2016unsupervised}.). \\
\cline{2-3}          & SPL (2020) \cite{wang2020unsupervised} &  Supervised locality preserving projection and selective pseudo-labeling strategy  \\
\cline{2-3}          & MDD (2020) \cite{jiang2020implicit} & Within-domain class imbalance and between-domain class distribution shift. \\
\cline{2-3}          & HoMM (2020) \cite{chen2020homm} & Higher-order moment matching for UDA. \\
\cline{2-3}          & GSDA (2020) \cite{hu2020unsupervised} & Model the relationship among the local distribution pieces and global distribution synchronously. \\
\cline{2-3}          & ETD(2020) \cite{li2020enhanced} & Attention mecanism for samples similarity  andattention scores for the transport distances. \\
\cline{2-3}          & BAIT (2020) \cite{yang2020unsupervised} & Source-free unsupervised domain adaptation. \\
\cline{2-3}          & DAEL (2021) \cite{zhou2021domain} & Ensemble Learning. \\
\cline{2-3}          & SHOT (2021) \cite{liang2020we} & Source-free unsupervised domain adaptation. \\
\cline{2-3}          & SHOT-plus (2021) \cite{liang2021source} & Source-free unsupervised domain adaptation. \\
\cline{2-3}          & SENTRY (2021) \cite{prabhu2021sentry} & Entropy Optimization. \\
\cline{2-3}          & RWOT (2021) \cite{xu2020reliable} & Shrinking Subspace Reliability  and weighted optimal transport strategy. \\
\cline{2-3}          & N2DC-EX (2021) \cite{tang2021nearest} & Source-free unsupervised domain adaptation. \\
    \hline
        \multirow{15}{*}{GAN} & Co-GAN (2016) \cite{liu2016coupled} & A group of GANs with partly weight sharing, discriminator and label predictor are unified. \\
\cline{2-3}          & DANN (2016)   \cite{ganin2016domain} & Domain classifier and label predictor. \\
\cline{2-3}          & UNIT (2017) \cite{liu2017unsupervised} & Use variational autoencoder as feature extractor \\
\cline{2-3}          & ADDA(2017) \cite{tzeng2017adversarial}  & Generalization  of Co-GAN \cite{liu2016coupled}. \\
\cline{2-3}          & PixelDA (2017) \cite{bousmalis2017unsupervised} & Generate instances follow target distribution with source samples. \\
\cline{2-3}          & GenToAdapt (2018) \cite{sankaranarayanan2018generate} & Two classifiers and one encoder to embed the instances into vectors. \\
\cline{2-3}          & SimNet (2018) \cite{pinheiro2018unsupervised} & Similarity-based classifier . \\
\cline{2-3}          & MADA (2018) \cite{pei2018multi} & Multi-domains. \\
\cline{2-3}          & DIFA (2018) \cite{volpi2018adversarial} & Extended ADDA  \cite{tzeng2017adversarial}  uses a pair of feature extractors. \\
\cline{2-3}          & CyCADA (2018) \cite{hoffman2018cycada} & Semantic consistency at both the pixel-level and feature-level. \\
\cline{2-3}          & SymNet (2019) \cite{zhang2019domain} & Category-level and domain-level confusion losses. \\
\cline{2-3}          & M-ADDA (2020) \cite{laradji2020m} & Triplet loss function and ADDA  \cite{tzeng2017adversarial}. \\
\cline{2-3}          & IIMT (2020) \cite{yan2020improve} & Mixup formulation and a feature-level consistency regularizer. \\
\cline{2-3}          & MA-UDASD (2020) \cite{li2020model} & Source-free unsupervised domain adaptation. \\
\cline{2-3}          & DM-ADA (2020) \cite{xu2020adversarial} & Domain mixup is jointly conducted on pixel and feature level. \\
    \hline
    \end{tabular}%
  \label{tab:addlabel1}%
\end{table*}%

\subsection{GAN-based}
\par \emph{DNN-based} UDA methods mainly focus on an appropriate measurement for the source and target domains. By contrast, \emph{GAN-based} UDA methods use the discriminator to fit this measurement function. Usually, in \emph{GAN-based} UDA methods, the generator ${\phi}_g$ is used to produce data followed by one distribution from another distribution, and the discriminator ${\phi}_d$ is used to judge whether the data generated follow the distribution of the target domain. Traditional GAN can not satisfy the demands to project two domains into the same space, so different frameworks based on GAN are proposed to cope with this challenge.
\par
\begin{figure}[!t]
\centering
\includegraphics[scale=0.37]{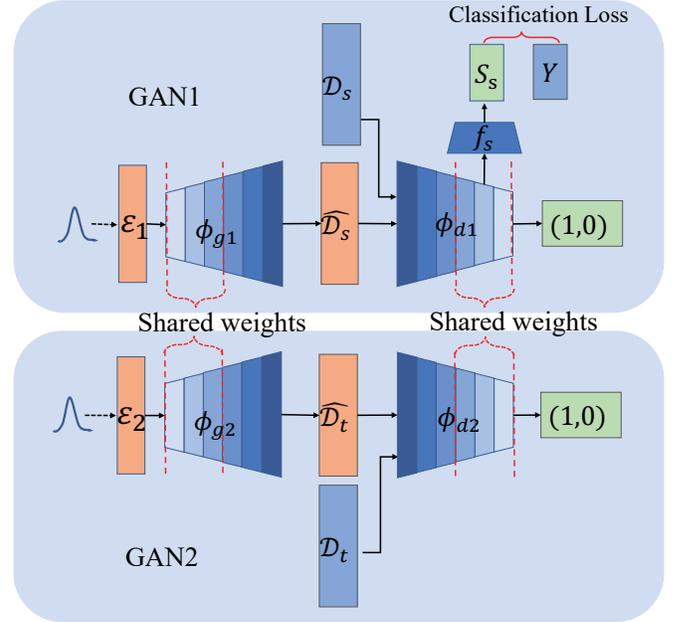}
\caption{The data stream framework of Co-GAN applicated in UDA. It consists of a pair of GANs: GAN1 and GAN2. GAN1 and GAN2 share the weight in the first layers of ${\phi}_g$ and last layers of ${\phi}_d$. $\mathcal{D}_s$ is the source domain. $\mathcal{D}_t$ is the target domain. ${\phi}_d$. $\widehat{\mathcal{D}_s}$ and ${\phi}_d$. $\widehat{\mathcal{D}_t}$ are generated by the noise. The first layers of ${\phi}_g$ is responsible for decoding high-level semantics and the last layers of ${\phi}_d$ is responsible for encoding high-level semantics. Add weight sharing constraint in these layers can guarantee similar high-level semantic representations of both domains with different low-level feature representations. 
}
\label{fig:co-gan}
\end{figure}

 In 2016, domain-adversarial neural network (DANN) \cite{ganin2016domain} and coupled generative adversarial networks (Co-GAN)\cite{liu2016coupled} are proposed to introduce adversarial learning into transfer learning. DANN uses a discriminator to ensure the feature distributions over the two domains are made similar. CO-GAN applies generator and discriminator all in UDA methods. It consists of a group of GANs, each corresponding to a domain. In UDA, there are two domains. The framework of CO-GAN is shown in Fig.  \ref{fig:co-gan}.

\par In deep transfer learning, we need to find the proper layers for MMD or weight sharing. In general, we could see that the networks which want to transfer the knowledge through domain adaption must pay more attention to the layers which are responsible for high-level semantic layers. In DAN, the first layers are for basic features and the high layers for semantic information are zoomed in where the last layers are chosen to be projected with MMD. In Co-GAN, also the semantic layers are chosen as the transferring layers (take notice, the first layers of DAN are not transferring layers between two domains, as it is transferring the feature extracting power of a mutual network to our domains' feature extracting part). The weight-sharing constraint in the first layers of the generator urges two instances from a different domain to extract the same semantics and are destructed into different low-level details in the last layers of ${\phi}_g$. In opposite, the discriminator learns the features from low-level to high-level, so if we add weight-sharing constraint in the last layers, this can stimulate it to learn a joint distribution of multi-domain images from different low-level representations.

Co-GAN contributed significant thought to UDA. Adversarial methods in domain adaptation have sprung up. For the job that relies on the synthesized instances to assist the domain adaptation process, they always perform not very well on real images such as the $OFFICE$ dataset. GenToAdapt-GAN\cite{sankaranarayanan2018generate} is proposed in cases where data generation is hard, even though the generator network they use performs a mere style transfer, yet this is sufficient for providing good gradient information for successfully aligning the domains. Unlike Co-GAN, there is just one generator and one discriminator. Additionally, there are two classifiers and one encoder to embed the instances into vectors.


\par Co-GAN and GenToAdapt adopt different strategies to train a classifier for an unlabeled domain. The biggest difference between Co-GAN and GenToAdapt-GAN is whether the feature extractor is the same. The feature extractor of Co-GAN is the GAN itself, but the feature extractor of GenToAdapt-GAN is a specialized encoder. In Co-GAN, GAN must do the jobs of adversarial process and encoding at the same time, but in GenToAdapt-GAN, these two jobs are separated which means GenToAdapt-GAN will be stabler and perform better when the data is complex. Most of the methods proposed in recent years are based on these two ways. \cite{liu2017unsupervised} adopted different GAN for different domains and weight-sharing. The main change is that the generator is replaced by VAE. ADDA (adversarial discriminative domain adaptation)\cite{tzeng2017adversarial} adopted the discriminative model as the feature extractor is based on Co-GAN. ADDA can be viewed as generalization  of CO-GAN framework. \cite{volpi2018adversarial}  extended ADDA using a pair of feature extractors. \cite{laradji2020m} uses a metric learning approach to train the source model on the source dataset by optimizing the triplet loss function as an optimized method and then using ADDA to complete its transferring process. SymNet \cite{zhang2019domain} proposed a two-level domain confusion scheme that includes category-level and domain-level confusion losses.
With the same feature extractor of the source and target domain,
MADA (multi-adversarial domain adaptation) \cite{pei2018multi} sets the generator as its feature extractor expanding the UDA problem to multi-domains. Similarity-based domain adaption network (SimNet)\cite{pinheiro2018unsupervised} uses discriminator as a feature extractor and a similarity-based classifier which compares the embedding of an unlabeled image with a set of labeled prototypes to classify an image. \cite{yan2020improve} using mixup formulation and a feature-level consistency regularizer to address the generalization performance for target data. \cite{xu2020adversarial} uses domain mixup on both pixel and feature level to improve the robustness of models.

There is also a very straightforward way to transfer the knowledge between domains: Generate new instances for the target domain. If we transfer the instance from the source domain into a new instance that followed a joint distribution of both domain and are labeled the same as its mother source instance, then we get a batch of ``labeled fake instances in target domain''. Training a classifier with these fake instances should be applicative to the real target data. In this way, we can easily use all the unsupervised adversarial domain adaptation methods in UDA as an effective data augmentation method. This accessible method also performs well in the deep clustering problem and is called pixel-level transfer learning.
\par
\begin{figure}[!t]
\centering
\includegraphics[scale=0.4]{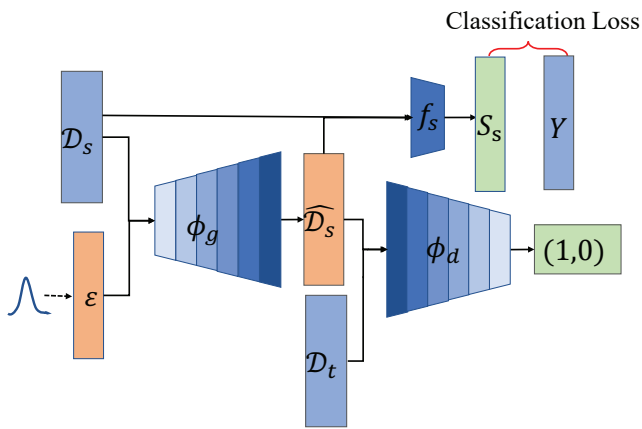}
\caption{An overview of the model architecture. The generator ${\phi}_g$ generates an image conditioned on a synthetic image which fed into discriminator as fake data and a noise vector $\varepsilon$. The discriminator ${\phi}_d$ discriminates between real and fake images. $\mathcal{D}_s$ is the source domain. $\mathcal{D}_t$ is the target domain. $\widehat{\mathcal{D}_s}$ is the fake image, $f_s$ is trained with generated data and source data. $Y$ means the real labels and $S_s$ denotes the predicted results.}
\label{fig:Pixel_GAN}
\end{figure}

 Unsupervised pixel–level domain adaptation with generative adversarial networks (Pixel-GAN) \cite{bousmalis2017unsupervised} aims at changing the images from the source domain to appear as if they were sampled from the target domain while maintaining their original content (label). The authors proposed a novel GAN-based architecture that can learn such a transformation in an unsupervised manner. The training process of Pixel-GAN is shown in Fig.~\ref{fig:Pixel_GAN}. It uses a generator ${\phi}_g$ to propose a fake image with the input composed of a labeled source image and a noise vector. The fake images will be discriminated against with target data by a discriminator ${\phi}_d$. At the same time, fake images $
\widehat{\mathcal{D}_s}
$
 and source images are put into a classifier $f_s$, when the model is convergent, the classifier can be used on the target domain.

\par On the whole, Pixel-GAN is a very explicit model, but this net relies on the quality of the generated images too much. Although the classifier can guarantee the invariant information of classes, it is also hard to perform on complex images. Pixel-level transferring and feature-level transferring are not going against each other, as pixel-level can transfer visual features and feature-level transferring can transfer the nature information of the instances. Cycle-Consistent adversarial domain adaptation (CyCADA) \cite{hoffman2018cycada} adapts representations at both the pixel-level and feature-level while enforcing semantic consistency. The authors enforce both structural and semantic consistency during adaptation using a cycle-consistency loss and semantics losses based on a particular visual recognition task. The semantics losses both guide the overall representation to be discriminative and enforce semantic consistency before and after mapping between domains.
Except for GAN, adopting data augmentation to transfer learning can also use traditional ways. \cite{sun2019unsupervised} provides the efficiency to make data augmentation in the target domain even it is unlabeled. It adds self-supervised tasks to target data and shows good performance. More important is that this skill can be combined with other domain adaptation methods such as CyCADA and DAN.

 \section {Future Directions of Deep Clustering}

\par Based on the aforementioned literature review and analysis, deep clustering has been applied to several domains, and we attach importance to several aspects worth studying further:
\begin{itemize}
    \item Theoretical exploration
\end{itemize}

\par Although remarkable  clustering performance has been achieved by designing even more sophisticated deep clustering pipelines for specific problem-solving needs, there is still no reliable theoretical analysis on how to qualitatively analyze the influence of feature extraction and clustering loss on final clustering. So, exploring the theoretical basis of deep clustering optimization is of great significance for guiding further research in this field.

\begin{itemize}
    \item Massive complex data processing
\end{itemize}
\par Due to the complexity brought by massive data, most of the existing deep clustering models are designed for specific data sets. Complex data from different sources and forms bring more uncertainties and challenges to clustering. At present, deep learning and graph learning are needed to solve complex data processing problems.

\begin{itemize}
    \item Model efficiency
\end{itemize}
\par Deep clustering algorithm requires a large number of samples for training. Therefore, in small sample data sets, deep clustering is prone to overfitting, which leads to the decrease of clustering effect and the reduction of the generalization performance of the model. On the other hand, the deep clustering algorithm with large-scale data has high computational complexity, so the model structure optimization and model compression technology can be adopted to reduce the computational load of the model and improve the efficiency in practical application conditions.

\begin{itemize}
    \item Fusion of multi-view data
\end{itemize}
\par In practical application scenarios, clustering is often not just with a single image information, but also available text and voice information. However, most of the current deep clustering algorithms can only use one kind of information and can not make good use of the existing information. The subsequent research can consider to fully integrate the information of two or more views and make full use of the consistency and complementarity of data of different views to improve the clustering effect. Furthermore, how to combine features of different views while filtering noise to ensure better view quality needs to be solved.

\begin{itemize}
    \item Deep clustering based on graph learning
\end{itemize}
\par In reality, a large number of data sets are stored in the form of graph structures. Graph structure can represent the structural association information between sample points. How to effectively use the structural information is particularly important to improve the clustering performance. Whether it is a single-view deep clustering or a relatively wide application of multi-view deep clustering, existing clustering methods based on graph learning still have some problems, such as the graph structure information is not fully utilized, the differences and importance of different views are not fully considered. Therefore, the effective analysis of complex graph structure information, especially the rational use of graph structure information to complete the clustering task, needs further exploration.

 \section {Summary of Deep Clustering Methods}
In this paper, we introduce recent advances in the field of deep clustering. This is mainly kind of data structures: single-view, semi-supervised, multi-view, and transfer learning. Single-view methods are the most important part of our survey, which inherits the problem settings of traditional clustering methods. We introduce them from the network they are based on. Among these networks, \emph{DAE-based} methods and \emph{DNN-based} methods are proposed earlier but limited with their poor performance in a real dataset. Compared to \emph{DAE-based} and \emph{CNN-based} methods, \emph{VAE-based} and \emph{GAN-based} methods attract attention in recent years for their strong feature extraction and sample generation power. Graph neural network is one of the most popular networks recently, especially in community discovery problems. So we also summarize the \emph{GNN-based} clustering methods. With the development of the internet, the data for clustering have different application scenarios, so we summarize some clustering methods which have different problem settings. Semi-supervised clustering methods cluster the data with constraints that can be developed from single-view clustering methods by adding a constraints loss. Multi-view clustering methods use the information of different views as a supplement. It has been used widely in both traditional neural networks and graph neural networks. Transfer learning can transfer the knowledge of a labeled domain to an unlabeled domain. We introduce clustering methods based on transfer learning with two types of networks: DNN and GAN. \emph{DNN-based} methods focus on the measurement strategy of two domains. \emph{GAN-based} methods use discriminators to fit the measurement strategy.

In general, single-view clustering has a long history and it is still a challenge especially on complex data. But the information outside should also be considered in application scenarios. For instance, the news reported by multiple news organizations; sensor signals decompose in the time and frequency domains; a mature dog classification network is useful to class the cats' images. Semi-supervised models, multi-view models, and unsupervised domain adaptation models consider multi-source information would attract more attention in practical application.

\ifCLASSOPTIONcaptionsoff
  \newpage
\fi

\bibliographystyle{unsrt}
\bibliography{0mybibfile}

\end{document}